\documentclass[default]{sn-jnl}

\usepackage{graphicx}%
\usepackage{multirow}%
\usepackage{amsmath,amssymb,amsfonts}%
\usepackage{amsthm}%
\usepackage{mathrsfs}%
\usepackage{tabularx}
\usepackage[linesnumbered, ruled, vlined]{algorithm2e}
\usepackage[title]{appendix}%
\usepackage{xcolor}%
\usepackage{textcomp}%
\usepackage{manyfoot}%
\usepackage{booktabs}%
\usepackage{listings}
\usepackage{graphicx}
\usepackage{color}
\usepackage{tabularx}
\usepackage{tabulary}
\usepackage{longtable}
\usepackage{booktabs} 
\usepackage{siunitx} 
\usepackage{caption}
\usepackage{booktabs}
\usepackage{siunitx}
\usepackage{makecell}
\usepackage{subcaption}
\usepackage{url}
\usepackage{geometry}
\usepackage{setspace} 
\usepackage{lineno}

\begin{document}

\title{An Explainable Vision Transformer with Transfer Learning Based Efficient Drought Stress Identification}

\author[1,2]{\fnm{Aswini Kumar} \sur{Patra}}\email{aswinipatra@gmail.com / akp@nerist.ac.in}

\author[3]{\fnm{Ankit} \sur{Varshney}} 

\author*[2]{\fnm{Lingaraj} \sur{Sahoo}}\email{ls@iitg.ac.in}

\affil[1]{\orgdiv{Dept. of Computer Science \& Engineering}, \orgname{North Eastern Regional Institute of Science and Technology}, \country{India}}

\affil[2]{\orgdiv{Dept. of Bio-Science and Bio-Engineering}, \orgname{Indian Institute of Technology Guwahati}, \country{India}}

\affil[3]{\orgdiv{Dept. of Physics}, \orgname{Indian Institute of Technology Guwahati}, \country{India}}

\abstract{Early detection of drought stress is critical for taking timely measures for reducing crop loss before the drought impact becomes irreversible. The subtle phenotypical and physiological changes in response to drought stress are captured by non-invasive imaging techniques and these imaging data serve as valuable resource for machine learning methods to identify drought stress. While convolutional neural networks (CNNs) are in wide use, vision transformers (ViTs) present a promising alternative in capturing long-range dependencies and intricate spatial relationships, thereby enhancing the detection of subtle indicators of drought stress. We propose an explainable deep learning pipeline that leverages the power of ViTs for drought stress detection in potato crops using aerial imagery. We applied two distinct approaches: a synergistic combination of ViT and support vector machine (SVM), where ViT extracts intricate spatial features from aerial images, and SVM classifies the crops as stressed or healthy and an end-to-end approach using a dedicated classification layer within ViT to directly detect drought stress. Our key findings explain the ViT model's decision-making process by visualizing attention maps. These maps highlight the specific spatial features within the aerial images that the ViT model focuses as the drought stress signature.  Our findings demonstrate that the proposed methods not only achieve high accuracy in drought stress identification but also shedding light on the diverse subtle plant features associated  with drought stress. This offers a robust and interpretable solution for drought stress monitoring for farmers to undertake informed decisions for improved crop management.}

\keywords{Stress Pheno-typing, Drought Stress, Machine Learning, Deep Learning, Vision Transformer, Support Vector Machine}



\maketitle

\section{Introduction}
Drought stress adversely affects the growth, development and yield of plants. The rise in global temperature coupled with the growing demand for water severely depletes soil water reserves. Water deficit and temperature stress can cause damage to the reproductive growth phase of crops, resulting in a substantial reduction in yield \cite{seleiman_drought_2021}. Furthermore, prolonged exposure to water deficit stress particularly in the reproductive stage can significantly affect grain quality. Ensuring sustainable crop yields in the face of escalating drought conditions has become a great challenge \cite{dinneny_developmental_2019}. The plant's response to drought stress is influenced by a number of factors such as stress duration, severity, genotype and its developmental stage \cite{yang_response_2021}. In plants, drought stress in early phase induces reduced stomatal conductance, transpiration and net carbon assimilation, whereas mid- and late-phases result in visible leaf senescence, stagnant growth and wilting. The early detection of drought induced phenotypic changes in plants, facilitates in preventing the adverse impacts of drought from irreversible damage through timely irrigation and other management practices. The point based, contact type and destructive methods based estimation of drought stress impact on crops are limited by good approximation of representative area as well as sample size and hence, difficult to implement over larger agricultural production systems. On the contrary, image-based, non-invasive and high-throughput analyses of subtle phenotypical changes in plants due to drought and other stresses, provides an efficient means of monitoring  stress levels in crops and good prediction of drought stress impact \cite{furbank_phenomics_2011}. Various imaging techniques and sensors are utilized to precisely capture stress responses, including simple red-blue-green (RGB) , thermal, multispectral, hyperspectral, or fluorescence imaging methods \cite{Tamimi_2022}. However, traditional image processing generates significant amount of variation in light intensity, shading, occlusion, etc. leading to variation in quality thereby increases complexity of image processing \cite{singh_deep_2018}. Machine learning (ML) techniques, on the other hand, solve such complex problems which are nonlinear in nature by identifying key feature to be extracted from the acquired image in real time, based on knowledge and/or domain expertise of the user. Hence, ML aids in better decision-making and informed actions in real-world scenarios with minimal human intervention \cite{liakos_machine_2018}. Deep learning (DL) \cite{arya_deep_2022} techniques however, help in extracting different information from the raw image data during input training, by means of various convolutions which allows larger learning capabilities, that result in the best classification/regression and higher performance and precision. Convolutional Neural Networks (CNNs) have been developed for providing a much-detailed representation of the image features, for a more discriminative and detailed analysis of plant images, leading to highly accurate classifications \cite{jiang_convolutional_2020}. Many studies have used deep CNN models to detect drought stress in affected plants, mostly solved either by finely tuned CNNs or by training from scratch. 

 Early work by Zhuang et al. \cite{zhuang_early_2017} utilized a combination of segmentation, color, and texture feature extraction, employing a gradient boosting decision tree (GBDT) method to detect water stress in maize. However, a subsequent study using the same RGB dataset demonstrated that a deep convolutional neural network (DCNN) significantly outperformed GBDT in terms of performance \cite{an_identification_2019}.
Sankararao et al. \cite{sankararao_machine_2023} developed a comprehensive pipeline for detecting water stress in groundnut canopies using hyperspectral imaging. The pipeline included steps for image quality assessment, denoising, and band selection, followed by classification using Support Vector Machine (SVM), Random Forest (RF), and Extreme Gradient Boosting (XGBoost/XGB). Transfer learning with a pre-trained DenseNet-121 model has been utilized to classify soybean color images into four levels of drought severity \cite{ramos-giraldo_drought_2020}. Similarly, a controlled dataset is created for chickpea crops to identify water stress at different stages—control, young seedling, and pre-flowering—using a CNN-LSTM hybrid model where CNNs that served as feature extractors and LSTM models for predicting water stress levels \cite{azimi_intelligent_2021}.
The evaluation of various CNN models, including AlexNet, GoogLeNet, and Inception V3, for distinguishing between stress and non-stress conditions in water-sensitive crops such as maize, okra, and soybean, revealed that GoogLeNet achieved the highest identification accuracy \cite{chandel_identifying_2021}.
Chlorophyll fluorescence images of wheat canopies have been studied employing a multi-step process involving segmentation and feature extraction with a correlation-based gray-level co-occurrence matrix (CGLCM) and color features, followed by classification using different ML classifiers. The tree-based methods, particularly random forest (RF) and extra trees classifier, is found giving superior performance \cite{gupta_drought_2023}.
A dataset of aerial images of potato canopies has been studied using a deep learning-based model for drought stress identification, leveraging various imaging modalities and their combinations \cite{butte_potato_2021}. An explainable deep learning framework was later proposed using the same dataset, based on CNN architectures and transfer learning, achieving improved accuracy \cite{patra_explainable_2024}.
A regression approach is applied to predict the drought tolerance coefficient using hyperspectral images of tea canopies; the comparison of SVM, RF, and Partial Least Squares Regression (PLSR) revealed that SVM was the most effective \cite{chen_hyperspectral_2022}. Machine learning and deep learning methods were compared, including DNN, SVM, and RF, for drought stress detection using full spectra and first-order derivative spectra \cite{dao_plant_2021}. Goyal et al. \cite{goyal_deep_2024} created a dataset of RGB images of maize crops and proposed a custom-designed CNN model that outperformed existing state-of-the-art CNN architectures in early drought stress detection.  However, Vision Transformers (ViTs) offer notable advantages over CNNs in capturing long-range dependencies and global context due to their self-attention mechanism, whereas CNNs rely on local receptive fields and often struggle to capture relationships across various parts of an image \cite{dosovitskiy_image_2021}. Additionally, ViTs exhibit greater flexibility with input image sizes and handle complex patterns and data variations more effectively \cite{chen_transunet_2021}.

The self-attention mechanism in ViTs enables more accurate classification by considering the entire image context, which enhances accuracy and robustness compared to traditional CNNs \cite{liu_swin_2021}. For example, Dosovitskiy et al. demonstrated that ViTs could outperform CNNs in image classification tasks, highlighting their potential for diversified applications \cite{dosovitskiy_image_2021}. Recent studies have effectively applied ViTs both in customized form \cite{bhowmik_customised_2024, singh_effective_2024} and CNN- ViT hybrid form to plant disease identification \cite{borhani_deep_2022}. Thakur et al. \cite{ thakur_vision_2023} developed a lightweight model that combines convolutional blocks from VGG 16 and Inception V7 with transformer components such as multi-head attention and multi-layer perceptron to effectively identify a wide range of plant diseases across multiple crops. The model leverages the local feature extraction capabilities of CNNs and the global feature modeling strength of vision transformers, enabling simultaneous extraction of both local and global features from images.
ViT has outperformed Inception V3 in terms of accuracy when distinguishing among nine different tomato leaf disease classes \cite{ barman_vit-smartagri_2024}. Parez et al. fine-tuned the vision transformer with a four-fold reduction in training parameters, achieving higher accuracy compared to CNN models for disease identification across three datasets \cite { parez_visual_2023}. Yu et al. \cite { yu_inception_2023} proposed a framework that integrates soft split token embedding and depth-wise convolutional modules into the vision transformer architecture, resulting in improved accuracy. Replacing the MLP module in the ViT encoder block with an Inception module improved accuracy in multi-crop disease classification while reducing the number of trainable parameters \cite{gole_trincnet_2023}. Thai et al. optimized the ViT architecture for cassava leaf disease detection by pruning less important attention heads and using sparse matrix operations, achieving a 2\% improvement in F1-score along with reduced model size and training costs \cite{thai_formerleaf_2023}. Hemalatha et al. developed a plant disease localization and classification model which uses co-scale, co-attention, and cross-attention mechanisms with a vision transformer in a multi-task learning framework \cite{hemalatha_multitask_2024}. Li et al. integrated a convolutional block attention module into the standard ViT encoder, enabling the network to filter out irrelevant information and focus on essential features, leading to improved crop disease classification in rice, wheat, and coffee \cite{li_pmvt_2023}.
Vallabhajosyula et al. \cite{ vallabhajosyula_novel_2024} proposed a novel framework that combines a transformer encoder with ResNet9 for plant disease classification, outperforming several classical CNN-based models. From the extensive literature review, we observe that while most existing works focus primarily on improving accuracy, few address the reduction of trainable parameters, and none explore the explainability of transformers. Our proposed work differs in two key aspects: first, by deciphering the attention mechanism, we emphasize model explainability; second, we devised ViT and support vector machine (SVM) combined framework and investigated its performance.

In this study, we devised a Vision Transformer (ViT)-based framework and fine-tuned it specifically for drought stress identification, achieving improved accuracy in detecting stress conditions. To demonstrate the model’s interpretability, we employed the inherent self-attention properties of Vision Transformers to produce attention maps. These maps offer meaningful insights into how the model arrives at its decisions, thereby increasing both the transparency and trustworthiness of its predictions. Additionally, we proposed a hybrid ViT+SVM framework that combines the rich feature representation capabilities of ViTs with the strong classification performance of Support Vector Machines (SVMs), resulting in a more robust drought stress identification model. 
 
\section{Materials and Methods}
In this section, we begin by introducing the experimental dataset. Next, we present the drought stress classification model using two different approaches. In the first approach, we employ the vision transformer with transfer learning. In the second approach, we propose a framework that utilizes a vision transformer as a feature extractor, followed by the integration of an SVM as the classifier. Additionally, we investigate the interpretability of the model by generating and analyzing the attention maps. This comprehensive use of the ViT elucidates how the spatial features of drought stress can be precisely identified. Finally, we discuss the performance metrics for the proposed approaches.
\subsection{Preparing the Data}
The potato crop aerial images utilized in this study have been sourced from a publicly accessible dataset that encompasses multiple modalities \cite{potato_data, butte_potato_2021}. Collected from a field at the Aberdeen Research and Extension Center, University of Idaho, these images serve as valuable resources for training machine learning models dedicated to crop health assessment in precision agriculture applications. Acquired using a Parrot Sequoia multi-spectral camera mounted on a 3DR Solo drone, the dataset features an RGB sensor with a resolution of $4,608 \times 3,456$ pixels and four monochrome sensors capturing narrow bands of light wavelengths: green (550nm), red (660nm), red-edge (735nm), and near-infrared (790nm), each with a resolution of $1,280 \times 960$ pixels. The drone flew over the potato field at a low altitude of 3 meters, with the primary objective of capturing drought stress in Russet Burbank potato plants attributed to premature plant senescence.

\begin{figure}[h!]
    \centering
    \begin{subfigure}{0.45\textwidth}
        \centering
        \includegraphics[width=0.9\textwidth]{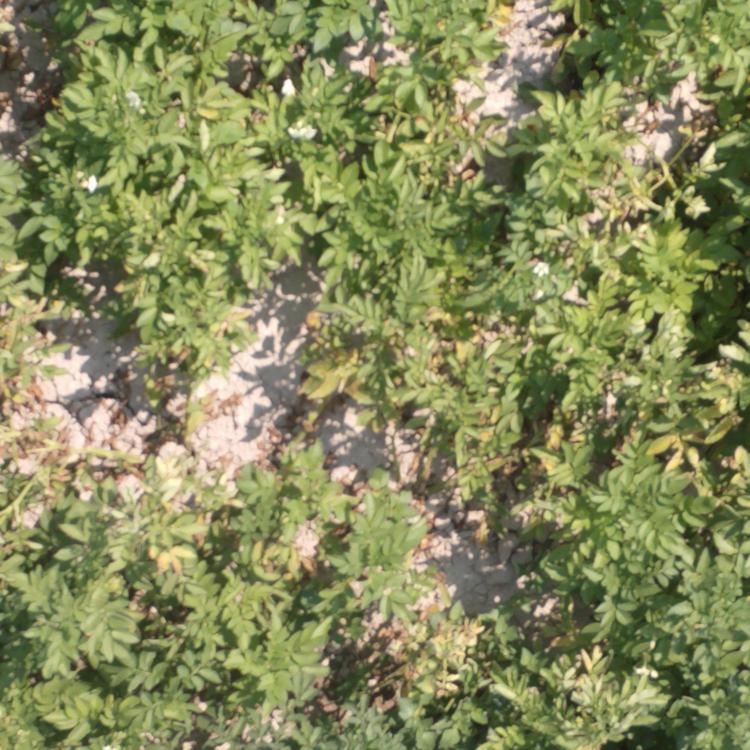}
        \caption{}
        \label{fig:rgb_sa}
    \end{subfigure}
    \begin{subfigure}{0.45\textwidth}
        \centering
        \includegraphics[width=0.9\textwidth]{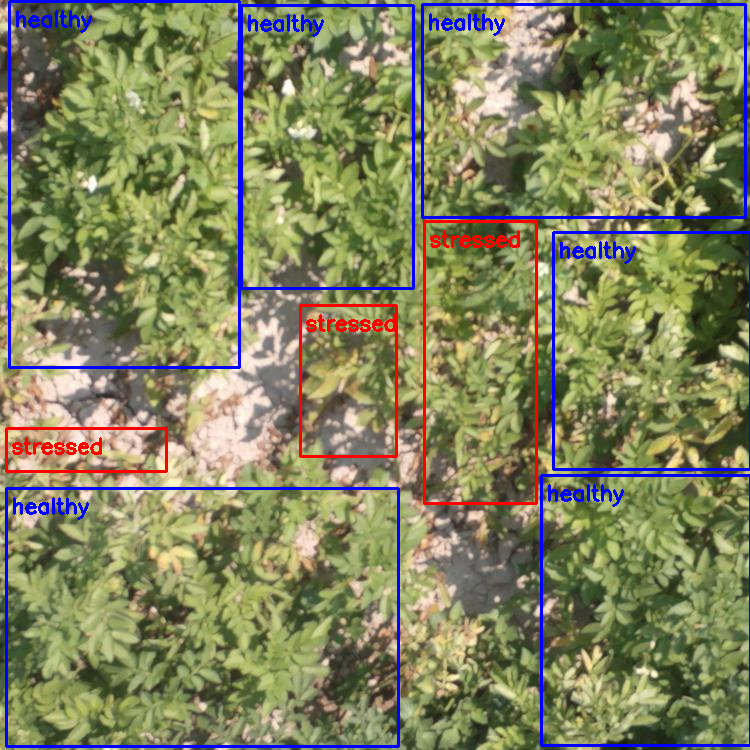}
        \caption{}
        \label{fig:hea_st}
    \end{subfigure}
    \caption{Field images showing \subref{fig:rgb_sa}) Sample RGB image and \subref{fig:hea_st}) Healthy and Stressed Labels.}
    \label{fig:sample_rgb}
\end{figure}

The dataset comprises of 360 RGB image patches in JPG format, each measuring 750×750 pixels. These patches were derived from high-resolution aerial images through cropping, rotating, and resizing operations. Data augmentation was applied to an initial set of 300 images as per procedure in Butte et al. \cite{butte_potato_2021}, expanding the training dataset to 1,500 images. The remaining 60 images were reserved exclusively for testing. No data augmentation was performed on the test images to ensure an unbiased evaluation. Training classification models requires labeled data with annotated regions of interest. In this study, the targets were regions containing healthy and stressed potato plants. These two conditions were visually distinguishable based on color—healthy plants appeared green, whereas stressed plants exhibited a yellowish hue. Manual annotation was performed using the open-source graphical tool LabelImg \cite{noauthor_labelimg_nodate}, allowing bounding boxes to be drawn around both healthy and stressed regions. The resulting annotations, including class labels and bounding box coordinates, were saved and used to generate ground truth data for training the proposed models.

Additionally, the dataset includes corresponding image patches from spectral sensors with red, green, red-edge, and near-infrared bands, each sized 416×416 pixels. However, we are only utilizing the RGB images due to the limitations of the low-resolution monochromatic images.

The augmented dataset, consisting of 1,500 images, was used for training, while the test set included 60 distinct images. From both training and test images, annotated windows (i.e., rectangular patches) were extracted based on the bounding box annotations. Each extracted window was labeled as either “healthy” or “stressed”, corresponding to the visual condition of the crop. As illustrated in Fig \ref{fig:sample_rgb}, the original image is shown in Fig \ref{fig:rgb_sa}, and the corresponding extracted windows are depicted in Fig \ref{fig:hea_st}. In this example, six windows represent healthy regions, while three represent stressed regions. Following this extraction process, the final training dataset comprised 11,915 stressed and 8,200 healthy image patches. From the 60 test images, we obtained 734 stressed and 401 healthy windows. The performance of the proposed model was first evaluated on this test set, and further validated through 5-fold cross-validation to ensure robustness and generalization.

\subsection{Vision Transformer (ViT)} 
A vision transformer (ViT) is a type of neural network architecture that has revolutionized the field of computer vision \cite{han_survey_2023}. Unlike traditional convolutional neural networks (CNNs), which process images pixel by pixel, ViT processes the input images in a sequential manner by dividing them into fixed-size patches, linearly embedding these patches, and then applying self-attention mechanisms for capturing global dependencies \cite{dosovitskiy_image_2021}. The VIT architecture used for our work is inspired by Dosovitskiy et al.\cite{dosovitskiy_image_2021} and depicted in Fig \ref{fig:vit_arc}. It processes images by dividing them into fixed-size embedded patches, linearly transforming these patches, and treating them as sequences. The transformer architecture is then applied, incorporating multi-head self-attention mechanisms \cite{parikh_decomposable_2016} that enable the model to capture long-range dependencies within the image. Layer normalization is applied before and after the multi-head attention, ensuring stable training, and a Multilayer Perceptron (MLP) head is added to the transformer's global representation for task-specific processing. 

\subsubsection{ViT Architecture} 
\begin{figure}[h!]
    \centering
    \begin{subfigure}{0.8\textwidth}
        \centering
        \includegraphics[scale=0.54]{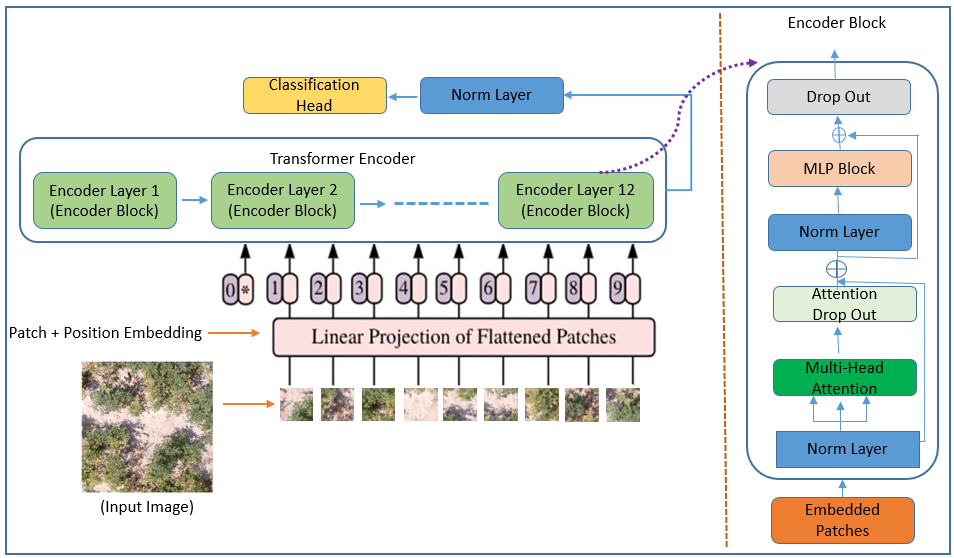}
        \caption{Vision Transformer Architecture}
        \label{fig:vit_arc}
    \end{subfigure}
    \vspace{0.5cm}
    \begin{subfigure}{0.8\textwidth}
        \centering
        \includegraphics[scale=0.52]{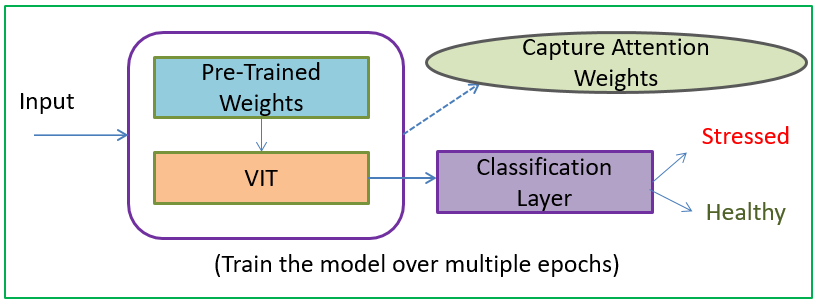}
        \caption{Vision Transformer with Transfer Learning}
        \label{fig:vit_tl}
    \end{subfigure}
    \vspace{0.5cm}
    \begin{subfigure}{0.8\textwidth}
        \centering
        \includegraphics[scale=0.51]{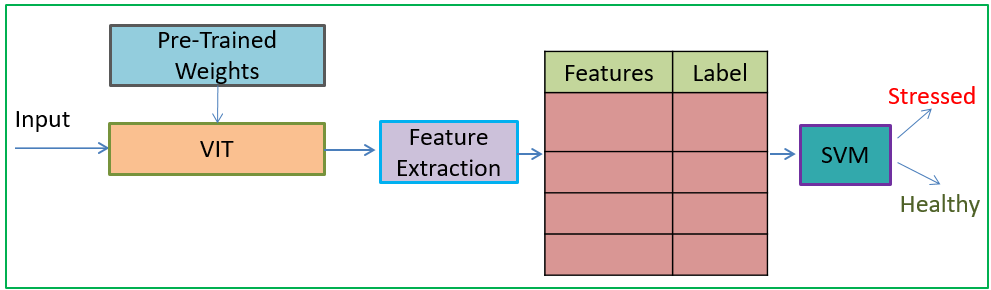}
        \caption{Integrating Vision Transformer and SVM}
        \label{fig:vit_svm}
    \end{subfigure}
    \caption{Vision Transformer based Approaches for Drought Stress Identification}
\end{figure}

The proposed ViT architecture is based on the layers in \cite{vaswani_attention_2023} and Dosovitskiy et al.\cite{dosovitskiy_image_2021} and comprises five main components: patch embedding, positional encoding, transformer encoder , normalization layer and a classification head. This is illustrated in Fig \ref{fig:vit_arc}.
\begin{itemize}
  \item \textbf{Patch Embedding}: Input images are divided into fixed-size patches, which are then linearly embedded to create a sequence of embeddings.
  
  \item \textbf{Positional Encoding}: To capture spatial information, positional encodings are added to the patch embeddings, allowing the model to understand the relative positions of different patches.
  
  \item \textbf{Transformer Encoder}: The embedded patches are fed into a Transformer Encoder, which is the core component of the ViT architecture. This encoder consists of twelve identical encoder layers stacked together. Each attention layer analyzes the relationships between pairs of patches, allowing the model to understand how different image regions interact and influence each other. Each attention layer consists of the following:
        \begin{itemize}
            \item Multi-Head Self-Attention: This mechanism allows the model to weigh the importance of different parts of the image. It captures global dependencies between the patches.
            \item MLP (Multi-Layer Perceptron) Block: This block introduces non-linearity to the network and further processes the information from the attention layer.
            \item Normalization Layers: Layer normalization is applied after the multi-head attention and MLP blocks to stabilize training.
            \item Dropout: The model uses dropout at two levels: within the attention mechanism and after the MLP block. Dropout is used to prevent over-fitting by randomly dropping out neurons during training.           
        \end{itemize}
   \item \textbf{Normalization Layer}: Layer normalization applied to the output of the encoder serves several crucial purposes: Reduces internal co-variate shift, improves gradient flow, acts as regularization, handles varying input distributions and accelerates convergence.     
    
  \item \textbf{Classification Head}: This head typically consists of a simple MLP layer that maps the feature representation to the desired output, such as class probabilities.
\end{itemize}

\subsubsection{Information Processing in ViT}
At the core of the Vision Transformer (ViT) is the concept of a token, which plays a crucial role in the model's ability to process and understand images. A token is a fixed-size vector that represents a small patch of the input image. These tokens are at the core of the Vision Transformer model, forming the input sequence to the Transformer layers. This token-based approach enables the model to process and understand the image by focusing on the relationships between these patches through self-attention. The detailed explanation is given below.

\begin{itemize}
    \item \textbf{Dividing the Image into Patches:}
    \begin{itemize}
        \item The input image is divided into smaller, fixed-size patches. For example, an image of size 224x224 pixels might be divided into patches of size 16x16 pixels, resulting in \(\left(\frac{224}{16}\right)^2 = 196\) patches.
    \end{itemize}
    
    \item \textbf{Flattening and Embedding:}
    \begin{itemize}
        \item Each image patch is then flattened into a one-dimensional vector. For instance, a 16x16 patch with 3 color channels (RGB) would be flattened into a vector of length \(16 \times 16 \times 3 = 768\).
        \item These flattened vectors (patch representations) are then linearly embedded into a higher-dimensional space. This is typically done using a learnable linear projection, transforming each vector into a fixed-size embedding, say of dimension 768.
    \end{itemize}
    
    \item \textbf{Tokens:}
    \begin{itemize}
        \item After linear embedding, each flattened and embedded patch becomes a "token." In this example, the image is transformed into a sequence of 196 tokens, each representing a 16x16 patch of the original image.
    \end{itemize}
    
    \item \textbf{Adding Positional Encoding:}
    \begin{itemize}
        \item Since the transformer model does not inherently understand the order or position of tokens, positional encodings are added to each token to incorporate information about its original position in the image. This helps the model understand spatial relationships between patches.
    \end{itemize}
    
    \item \textbf{Processing by Transformer Layers:}
    \begin{itemize}
        \item These tokens are then processed by the Transformer layers, which include self-attention mechanisms. The self-attention mechanism computes relationships between these tokens to understand how different parts of the image relate to one another.
    \end{itemize}
\end{itemize}

\subsection{ViT with Transfer Learning}
The proposed framework, as shown in Fig \ref{fig:vit_tl} effectively combines transfer learning, the power of the Vision Transformer and attention-based interpretability to address the challenging task of drought stress identification in potato crop images captured in natural settings. A core component of this approach is the utilization of pre-trained weights. This technique, known as \textit{transfer learning}, involves leveraging knowledge gained from solving one problem (often a large-scale image classification task) and applying it to a different but related problem. Specifically, we used the model initialized with pre-trained weights from training on the ImageNet-1k dataset with 1000 classes. By employing pre-trained weights, the model can benefit from the rich feature representations learned from a massive dataset, accelerating training and potentially improving performance. Using this the ViT is trained and fine-tuned over multiple epochs using a combination of \textit{binary cross-entropy} loss, \textit{Adam/AdamW} optimizer and learning rate tuning. Experimenting with different learning rates is essential to find the optimal value for convergence and generalization. Eventually, the classification layer is responsible for making the final prediction. It takes the output of the ViT model and maps it to two classes: "healthy" and "stressed." This layer typically consists of a fully connected neural network with a sigmoid activation function for binary classification.

To enhance model interpretability, the architecture incorporates a mechanism to capture attention weights. Attention weights reveal which parts of the input image the model focuses on when making a decision. By visualizing these weights, researchers can gain insights into the model's decision-making process and identify key image features that contribute to the classification. 

\subsubsection{Attention Maps} \label{sec:atten}


Attention maps provide insights into how the model focuses on different parts of the image during the self-attention mechanism. This mechanism allows the model to selectively attend to specific areas while processing visual information. The image is first divided into patches. Self-attention then helps the model prioritize relevant patches and their relationships for effective feature extraction.

Attention maps act as a visual representation of the weights assigned by the self-attention mechanism to each patch. These maps can be visualized for each self-attention layer within the ViT model, as each layer progressively learns more intricate relationships between these image features. Higher weights indicate a greater focus on a specific patch and its connection to others. By analyzing these maps, we can essentially see the model's "thought process" during image understanding. We can identify which regions it prioritizes for information extraction. The following section highlights the computation behind the attention mechanism \cite{vaswani_attention_2023}. 

At its core, self-attention computes a weighted sum of the values (features) based on the similarities (attention scores) between different positions in the sequence. This is achieved through three learnable matrices: Query (\(Q\)), Key (\(K\)), and Value (\(V\)). 

\textbf{Query Matrix ($Q$):} The query matrix is responsible for capturing the information about the current token being processed. It learns to encode the features of the token in a format suitable for comparison with other tokens. Each token in the sequence is associated with a query vector, which represents its characteristics in the context of the entire sequence.

\textbf{Key Matrix ($K$):} The key matrix holds information about the relationship between the current token and other tokens in the sequence. It learns to encode the features that determine how relevant each token is to the current token.

\textbf{Value Matrix ($V$): } 

The key matrix holds information about the relationship between the current token and other tokens in the sequence. It learns to encode the features that determine how relevant each token is to the current token.

Given a sequence of tokens \(X = [x_1, x_2, ..., x_n]\), the attention scores \(A\) are computed as:

\[ A = \text{softmax} \left( \frac{QK^T}{\sqrt{d_k}} \right) \]

where \( Q = XW_Q \), \( K = XW_K \), \( V = XW_V \), and \( W_Q, W_K, W_V \) are weight matrices. \( d_k \) represents the dimensionality of the key vectors.

Once the attention scores are computed, they are used to compute a weighted sum of the values:

\[ \text{Attention}(Q, K, V) = A \cdot V \]

where \( A \) is the attention matrix.








To capture different relationships between tokens, ViT (and Transformers in general) often employ multiple attention heads. Each head learns different sets of \(Q, K, V\) weight matrices and computes separate attention scores and weighted sums. The results from all heads are concatenated and linearly transformed to maintain a consistent output dimension.
Since self-attention does not inherently consider the order of tokens, positional encoding is typically added to the token embeddings to provide positional information.
 
\textbf{Algorithm for Capturing Attention Weights:}
To enhance model interpretability, the architecture incorporates a mechanism to capture attention weights. The algorithm \ref{alg:vit_binary} outlines the steps during the training of ViT to capture and utilize attention weights for the drought identification task. The key components of this class include the initialization, forward pass, attention weight capture, output size determination, and retrieval of attention weights. By visualizing these weights, researchers can gain insights into the model's decision-making process and identify key image features that contribute to the classification. 

\textbf{Initialization}
The VisionTransformerBinary class is initialized with a pre-trained Vision Transformer model passed as \texttt{vit\_model}. During initialization, the model assigns the provided \texttt{vit\_model} to its own \texttt{vit} attribute. Additionally, it sets up a fully connected (linear) layer (\texttt{fc}) with a size appropriate to the output of the Vision Transformer. This layer is responsible for converting the output of the Vision Transformer into a format suitable for binary classification. An empty list \texttt{attn\_weights} is also initialized to store the attention weights captured during the forward pass.

\textbf{Forward Pass}
The \texttt{forward} function is central to the operation of the VisionTransformerBinary class. When an input image \texttt{x} is passed through the model, the function first clears any previously stored attention weights. This ensures that the attention weights list is fresh and only contains data relevant to the current input.

Next, the function registers hooks on the self-attention layers of each block in the Vision Transformer. These hooks are responsible for capturing the attention weights during the forward pass. For each block in the Vision Transformer's encoder layers, the self-attention layer is accessed, and a hook is registered to capture its attention weights using the \texttt{\_capture\_attn\_weights} helper function. The hooks are stored in a list to facilitate their removal later.

With the hooks in place, the input image is passed through the Vision Transformer and then through the fully connected layer. This produces the model's output for the given input. After the forward pass is complete, the hooks are removed to free up memory, ensuring that they do not persist and interfere with future operations.

\textbf{Attention Weights Capture}
The \texttt{\_capture\_attn\_weights} helper function is designed to capture the attention weights during the forward pass. It is triggered by the hooks registered on the self-attention layers. This function receives the module, input, and output as arguments. From the input, it extracts the query, key, and value components, which are essential for computing the attention weights. These components, along with the output, are appended to the \texttt{attn\_weights} list. By capturing these components, the model can later compute and analyze the attention scores, which provide insights into the regions of the input image that the model focuses on during classification.

\textbf{Output Size Determination}
The \texttt{\_get\_output\_size} helper function determines the output size of the Vision Transformer model. This function performs a forward pass with a zero tensor of appropriate dimensions through the Vision Transformer. By doing so, it captures the shape of the output tensor produced by the Vision Transformer. The size of the last dimension of this output tensor is then returned, which is used to initialize the fully connected layer with the correct input size.

\textbf{Retrieval of Attention Weights}
The \texttt{get\_attention\_weights} function provides a simple interface to retrieve the captured attention weights. It returns the \texttt{attn\_weights} list, allowing external components or functions to access the attention weights for further analysis or visualization.

The modular design of the class, with separate functions for initialization, forward pass, attention weight capture, output size determination, and retrieval, provides a clear and maintainable structure. This design facilitates the integration of attention-based insights into the classification process, enhancing the interpretability and performance of the model. 

\begin{algorithm}[H]
\caption{Vision Transformer Class to capture Attention Weights}
\label{alg:vit_binary}

\KwIn{Pre-trained Vision Transformer model}
\KwOut{Modified Vision Transformer model with attention weights capture}

\SetKwFunction{FMain}{VisionTransformerBinary}
\SetKwProg{Fn}{Class}{:}{}
\Fn{\FMain{$vit\_model$}}{
    \KwData{$vit\_model$: Vision Transformer model}
    
    \tcp{Initialization}
    Initialize $vit$ with $vit\_model$\;
    Initialize $fc$ with a linear layer of appropriate output size\;
    Initialize $attn\_weights$ as an empty list\;
    
    \SetKwFunction{FForward}{forward}
    \SetKwProg{Fn}{Function}{:}{}
    \Fn{\FForward{$x$}}{
        \tcp{Clear previous attention weights}
        Clear $attn\_weights$\;
        
        \tcp{Register hooks to capture attention weights}
        Initialize $hooks$ as an empty list\;
        \For{each $block$ in $vit.encoder.layers$}{
            $attn\_layer \leftarrow block.self\_attention$\;
            $hook \leftarrow attn\_layer.register\_forward\_hook(\_capture\_attn\_weights)$\;
            Append $hook$ to $hooks$\;
        }
        
        \tcp{Pass input through the Vision Transformer}
        $x \leftarrow vit(x)$\;
        $x \leftarrow fc(x)$;
        
        \tcp{Remove hooks to free up memory}
        \For{each $hook$ in $hooks$}{
            Remove $hook$\;
        }
        
        \Return{$x$}\;
    }
    
    \SetKwFunction{FCaptureAttn}{\_capture\_attn\_weights}
    \SetKwProg{Fn}{Function}{:}{}
    \Fn{\FCaptureAttn{$module, input, output$}}{
        \tcp{Capture attention weights}
        Extract $query, key, value$ from $input$\;
        Append $(query, key, value, output)$ to $attn\_weights$\;
    }
    
    \SetKwFunction{FGetOutputSize}{\_get\_output\_size}
    \SetKwProg{Fn}{Function}{:}{}
    \Fn{\FGetOutputSize{}}{
        \tcp{Determine the output size of the Vision Transformer}
        Initialize $output$ with a forward pass of zeros through $vit$\;
        \Return{size of the last dimension of $output$}\;
    }
    
    \SetKwFunction{FGetAttnWeights}{get\_attention\_weights}
    \SetKwProg{Fn}{Function}{:}{}
    \Fn{\FGetAttnWeights{}}{
        \Return{$attn\_weights$}\;
    }
}
\end{algorithm}

\subsection{Integrating Vision Transformer (ViT) and Support Vector Machine (SVM)}

This framework focuses on combining a vision transformer (ViT)  and a support vector machine (SVM) within a three-step approach for effective stress identification.
\begin{itemize}
  \item Feature Extraction: For each image in the dataset, extract the final hidden state or pooled representation from the ViT model to obtain a feature vector. Let $\mathbf{X}$ represent the set of embedded features extracted from a dataset of images, and $\mathbf{y}$ represent the corresponding class labels.
  
  \item Training SVM: Train an SVM classifier using the extracted features $\mathbf{X}$ and their corresponding class labels $\mathbf{y}$.
  
  \item Classification: For a new, unseen image, extract features using the pre-trained ViT model and use the trained SVM classifier to predict its class.
\end{itemize}
\subsubsection{Support Vector Machine (SVM)}
Support Vector Machines \cite{hearst_support_1998} aim to find a hyperplane that best separates a given set of data points into different classes. Given a set of labeled data points $(\mathbf{x}_1, y_1), (\mathbf{x}_2, y_2), \ldots, (\mathbf{x}_n, y_n)$ where $\mathbf{x}_i$ is the feature vector of the $i$-th data point, and $y_i$ is its corresponding class label ($y_i \in \{1, 0\}$ for binary classification), SVM seeks to find a hyperplane defined by the equation:

\[
\mathbf{w} \cdot \mathbf{x} + b = 0
\]

where $\mathbf{w}$ is the weight vector and $b$ is the bias.

The goal is to maximize the margin, which is the distance between the hyperplane and the nearest data point from each class. The margin is computed as the perpendicular distance from a data point $\mathbf{x}_i$ to the hyperplane:

\[
\text{margin} = \frac{1}{\|\mathbf{w}\|} \cdot | \mathbf{w} \cdot \mathbf{x}_i + b |
\]

Subject to the constraint that for all data points:

\[
y_i (\mathbf{w} \cdot \mathbf{x}_i + b) \geq 1
\]

This constraint ensures that data points are correctly classified and lie on the correct side of the hyperplane.

The SVM optimization problem can be formulated as:

\[
\text{Minimize} \quad \frac{1}{2} \| \mathbf{w} \|^2
\]

Subject to the constraints:

\[
y_i (\mathbf{w} \cdot \mathbf{x}_i + b) \geq 1 \quad \text{for all } i
\]

This is the primal form of the optimization problem. However, the SVM problem is often reformulated in its dual form, which introduces Lagrange multipliers $\alpha_i$ to handle the constraints. The dual formulation is:

\[
\text{Maximize} \quad \sum_{i=1}^{n} \alpha_i - \frac{1}{2} \sum_{i=1}^{n} \sum_{j=1}^{n} \alpha_i \alpha_j y_i y_j \mathbf{x}_i \cdot \mathbf{x}_j
\]

Subject to the constraints:

\[
0 \leq \alpha_i \leq C \quad \text{for all } i
\]

\[
\sum_{i=1}^{n} \alpha_i y_i = 0
\]

The solution to the dual problem provides the values of $\alpha_i$, and the weight vector $\mathbf{w}$ and bias $b$ can be obtained from these values.

In cases where the data is not linearly separable, SVM can be extended to handle non-linear decision boundaries using the kernel trick. The feature space is implicitly mapped to a higher-dimensional space, making it possible to find a linear separating hyperplane in that space.

Classification is performed based on the decision function derived from the trained model. Once the SVM model is trained with a set of support vectors, it identifies a hyperplane that best separates different classes in the feature space. The decision function for a new data point $\mathbf{x}$ is given by:

\[ f(\mathbf{x}) = \mathbf{w} \cdot \mathbf{x} + b \]

Here, $\mathbf{w}$ is the weight vector, and $b$ is the bias term. The sign of $f(\mathbf{x})$ determines the predicted class:

\[
\begin{cases}
    \text{Class 1}, & \text{if } f(\mathbf{x}) > 0 \\
    \text{Class 0}, & \text{if } f(\mathbf{x}) < 0
\end{cases}
\]

The magnitude of $f(\mathbf{x})$ provides a measure of how far the data point is from the decision boundary. Larger magnitudes indicate greater confidence in the classification.

The key role of support vectors in this process is that they are the data points lying closest to the decision boundary. Support vectors effectively determine the position and orientation of the hyperplane. The optimization process in SVM aims to maximize the margin between the classes, and support vectors are the data points defining the edges of this margin. In practice, many data points do not significantly contribute to the definition of the decision boundary. Only the support vectors, with non-zero Lagrange multipliers $\alpha_i$ in the dual formulation, are crucial for determining the hyperplane. This property makes SVM memory-efficient and computationally faster, especially in high-dimensional spaces.`

\subsubsection{ViT+SVM Framework}
The process begins with input images that are fed into the ViT. The ViT, pre-trained on a vast dataset, is adept at extracting meaningful features from the images. Once the features are extracted by the ViT, they are compiled into a feature matrix, which also includes the corresponding labels indicating whether the plants in the images are healthy or stressed. This matrix forms the input to the SVM, a robust classifier known for its effectiveness in handling high-dimensional data. The SVM is trained to discern between the two classes—healthy and stressed—based on the features provided. Finally, the trained SVM predicts the class of new images, categorizing them as either healthy or stressed. The entire approach is depicted in Fig \ref{fig:vit_svm}.
The efficacy of this framework is evaluated using a designated test set and k-fold cross-validation. 
\subsection{Performance Evaluation Metrics}
The model's performance underwent assessment using various evaluation metrics, including accuracy, precision, recall (sensitivity) and the Receiver Operating Characteristic (ROC) curve. These metrics are computed based on the counts of true positives (TP), true negatives (TN), false positives (FP), and false negatives (FN), which collectively form a 2x2 matrix known as the confusion matrix. In this matrix, TP and TN indicate the accurate predictions of water-stressed and healthy potato crops, respectively. FP, termed as type 1 error, denotes predictions where the healthy class is inaccurately identified as water-stressed. FN, referred to as type 2 error, represents instances where water-stressed potato plants are incorrectly predicted as healthy. The classification accuracy is a measure of the ratio between correct predictions for both stressed and healthy images and the total number of images in the test set.

\[ \text{Accuracy} = \frac{\text{True Positive} + \text{True Negative}}{\text{Total Population}} \]

\[ \text{Precision} = \frac{\text{True Positive}}{\text{True Positive} + \text{False Positive}} \]

\[ \text{Recall} = \frac{\text{True Positive}}{\text{True Positive} + \text{False Negative}} \]

\[ \text{F1-score} = 2 \cdot \frac{\text{Precision} \cdot \text{Recall}}{\text{Precision} + \text{Recall}} \]

Cross-validation is crucial in model development for two key reasons: it helps prevent over-fitting by assessing a model's performance across different subsets of the data, and it ensures the model's generalization ability, providing a reliable estimation of its effectiveness under various conditions. The model was trained and evaluated using k-fold cross-validation, a robust technique for assessing the generalization performance of the model. In each iteration of the k-fold cross-validation process, the dataset was partitioned into $k$ folds, and the model was trained on $k-1$ folds while being validated on the remaining fold. This process was repeated k times, ensuring that every fold had the opportunity to serve as the validation set. For each fold, the Receiver Operating Characteristic (ROC) curve was plotted, illustrating the trade-off between true positive rate and false positive rate at various thresholds. After completing the k-fold cross-validation, the individual ROC curves were aggregated, and the mean ROC curve was calculated and plotted. AUC (area under the curve) measures the entire two-dimensional area underneath the entire ROC curve. AUC provides an aggregate measure of performance across all possible classification thresholds. One way of interpreting AUC is as the probability that the model ranks a random positive example more highly than a random negative example. AUC ranges in value from 0 to 1. A model whose predictions are 100\% wrong has an AUC of 0.0; one whose predictions are 100\% correct has an AUC of 1.0.This comprehensive approach provides a more reliable estimation of the model's performance, capturing its consistency across different subsets of the data and enhancing the overall assessment of its predictive capabilities.

\section{Results and Discussion}
In this section, we present the experimental results of our proposed model for identifying drought stress in potato crop field images. First, we distinguish between healthy and stressed images. Then, we identify the spatial features responsible for the stress. Our experiments with the proposed Vision Transformer (ViT) framework were conducted in two ways:
\begin{itemize}
    \item ViT with Transfer Learning (ViT-TL): Leveraging pre-trained weights.
    \item ViT+SVM with Optimal Weights. The optimal weights are the ones at which the model performs best while executing the ViT with pre-trained weights in the first case.
\end{itemize}

We used the \textit{PyTorch} library and its sub-packages to implement deep learning functionalities, particularly employing \textit{torch} and \textit{torch.nn} for tensor operations and neural network construction. For handling image data, we utilized \textit{Torchvision's} models for accessing pre-trained architectures and transforms for preprocessing, which included resizing images to (224, 224) pixels and converting them into tensors. Additionally, \textit{TQDM} was used to generate progress bars for better training visibility. For data analysis and pre-processing, we employed \textit{pandas}, \textit{NumPy}, and \textit{scikit-learn} for structured data manipulation, numerical computations, and machine learning utilities.

\subsection{Performance of ViT with Transfer Learning}
To adapt the Vision Transformer (ViT) architecture (as depicted in Fig \ref{fig:vit_arc}) for our specific task of binary classification, we began by configuring the model using the \textit{models.ViT-B/16 } variant. Subsequently, we loaded custom pre-trained weights into the ViT model to realize vision transformer with transfer learning approach as shown in Fig \ref{fig:vit_tl}. This step was crucial as it transferred learned representations from a previously trained model to our current architecture, leveraging prior knowledge to enhance performance. A custom class was designed (as shown in Algorithm \ref{alg:vit_binary}) to configure encoder layers as trainable or frozen, with methods to adjust various parameters. It also includes functionality to capture attention weights, crucial for analyzing the model's focus on specific image regions.

\begin{table}[hbt!]
\centering
\caption{Training parameters of the model under different scenarios.}
\label{tab:Training_Parameters}
\begin{tabular}{lcccccccc}
\toprule
\multirow{2}{*}{Scenario} & \multirow{2}{*}{Model} & \multirow{2}{*}{\shortstack{No. of \\ Trainable Layers}} & \multirow{2}{*}{\shortstack{Learning Rate \\ \& Optimizer}} & \multicolumn{2}{c}{Callback Parameters} & \multirow{2}{*}{\shortstack{Batch \\ Size}} & \multirow{2}{*}{\shortstack{Attention \\ Dropout}} & \multirow{2}{*}{\shortstack{MLP \\ Dropout}} \\
\cmidrule(lr){5-6}
 &  &  &  & Patience & Factor &  &  &  \\
\midrule
Scenario 1 & ViT-B/16  & Last encoder block & 0.001(Adam) & 5 & 0.2 & 128 & 0 & 0 \\
Scenario 2 & ViT-B/16  & Last two encoder blocks & 0.001(Adam) & 5 & 0.2 & 128 & 0 & 0 \\
Scenario 3 & ViT-B/16  & Last three encoder blocks & 0.001(AdamW) & 5 & 0.2 & 128 & 0 & 0 \\
Scenario 4 & ViT-L/16  & Last three encoder blocks & 0.001(AdamW) & 5 & 0.2 & 128 & 0 & 0 \\
Scenario 5 & ViT-B/16  & Last three encoder blocks & 0.001(AdamW) & 2 & 0.2 & 128 & 0 & 0 \\
Scenario 6 & ViT-B/16  & Last three encoder blocks & 0.001(AdamW) & 2 & 0.2 & 64 & 0 & 0 \\
Scenario 7 & ViT-B/16  & Last two encoder blocks & 0.001(AdamW) & 5 & 0.2 & 128 & 0.1 & 0.1 \\
Scenario 8 & ViT-B/16  & Last two encoder blocks & 0.001(AdamW) & 5 & 0.2 & 128 & 0.1 & 0.2 \\
Scenario 9 & ViT-B/16  & All encoder blocks & 0.001(AdamW) & 5 & 0.2 & 128 & 0 & 0 \\
Scenario 10 & ViT-B/16  & All encoder blocks & 0.001(AdamW) & 5 & 0.2 & 64 & 0 & 0 \\
Scenario 11 & ViT-B/16  & Last two encoder blocks & 0.001(Adam) & 5 & 0.2 & 128 & 0.1 & 0.2 \\
\bottomrule
\end{tabular}
\end{table}

Table~\ref{tab:Training_Parameters} presents the training configurations adopted across eleven experimental scenarios involving Vision Transformer (ViT)-based models. To balance accuracy with computational efficiency, we primarily fine-tuned a limited number of encoder layers within the pre-trained ViT models instead of retraining the entire architecture. Specifically, the number of trainable encoder blocks was varied across scenarios, starting with only the final block in Scenario 1 and gradually increasing up to all encoder blocks in Scenarios 9 and 10. The ViT-B/16 model, known for its relatively lightweight design, served as the backbone for most experiments. An exception was Scenario 4, where we employed the larger ViT-L/16 model, which has approximately four times the number of parameters compared to ViT-B/16. Despite its increased capacity, ViT-L/16 did not yield a notable improvement in accuracy, leading us to retain ViT-B/16 in subsequent scenarios for better scalability and efficiency. Key training elements were systematically varied to evaluate their effects. Both Adam and AdamW optimizers were tested with a fixed learning rate of 0.001, with AdamW being preferred in most cases due to its improved regularization capabilities. Callback settings included early stopping and learning rate reduction on plateau, controlled by a patience of 5 and a factor of 0.2. However, Scenarios 5 and 6 employed a reduced patience of 2 to accelerate convergence. Batch sizes were set to either 64 or 128 to investigate their influence on model convergence and generalization. Finally, to combat overfitting, additional attention and MLP dropout layers were integrated in Scenarios 7, 8, and 11.

\begin{table}[hbt!]
\centering
\caption{Model performance across different scenarios.}
\label{tab:Model_Performance}
\begin{tabular}{lcccccc}
\toprule
Scenario & \shortstack{Training \\ Accuracy} & \shortstack{Validation \\ Accuracy} & \shortstack{Training \\ Loss} & \shortstack{Validation \\ Loss} & \shortstack{Test \\ Accuracy} & \shortstack{Epoch \\ No.} \\
\midrule
Scenario 1  & 0.9899 & 0.9816 & 0.0272 & 0.0466 & 0.9039 & 16 \\
Scenario 2  & 0.9929 & 0.9831 & 0.0190 & 0.0563 & 0.9083 & 20 \\
Scenario 3  & 0.9939 & 0.9906 & 0.0178 & 0.0294 & 0.9057 & 16 \\
Scenario 4  & 0.9431 & 0.9443 & 0.1468 & 0.1433 & 0.8960 & 14 \\
Scenario 5  & 0.9935 & 0.9901 & 0.0191 & 0.0322 & 0.8819 & 17 \\
Scenario 6  & 0.9940 & 0.9876 & 0.0180 & 0.0370 & 0.8995 & 16 \\
Scenario 7  & 0.9833 & 0.9796 & 0.0421 & 0.0444 & 0.9127 & 18 \\
Scenario 8  & 0.9765 & 0.9747 & 0.0613 & 0.0876 & 0.9162 & 20 \\
Scenario 9  & 0.9377 & 0.9513 & 0.1589 & 0.1344 & 0.9119 & 16 \\
Scenario 10 & 0.9320 & 0.9274 & 0.1721 & 0.1747 & 0.9075 & 19 \\
Scenario 11 & 0.9707 & 0.9672 & 0.0780 & 0.0896 & 0.8942 & 16 \\
\bottomrule
\end{tabular}
\end{table}

Table \ref{tab:Model_Performance} summarizes the training and validation accuracy and loss for each scenario along with the final test accuracy and the epoch number at which early stopping was triggered. Across most scenarios, the training and validation accuracies exceed 97\%, demonstrating strong convergence. Scenario 5, despite high training accuracy (99.35\%), showed comparatively lower test accuracy (88.19\%), indicating potential overfitting. In contrast, Scenario 8 exhibited the highest generalization with a test accuracy of 91.62\%.

\begin{table}[hbt!]
\centering
\caption{Confusion matrix components and test accuracy across different scenarios.}
\label{tab:Confusion_Matrix_vit}
\begin{tabular}{lccccc}
\toprule
Scenario & TP & TN & FP & FN & \shortstack{Test \\ Accuracy} \\
\midrule
Scenario 1  & 647 & 379 & 22 & 87  & 0.9039 \\
Scenario 2  & 661 & 370 & 31 & 73  & 0.9083 \\
Scenario 3  & 650 & 378 & 23 & 84  & 0.9057 \\
Scenario 4  & 638 & 379 & 22 & 96  & 0.8960 \\
Scenario 5  & 617 & 384 & 17 & 117 & 0.8819 \\
Scenario 6  & 639 & 382 & 19 & 95  & 0.8995 \\
Scenario 7  & 663 & 373 & 28 & 71  & 0.9127 \\
Scenario 8  & 661 & 379 & 22 & 73  & 0.9162 \\
Scenario 9  & 645 & 390 & 11 & 89  & 0.9119 \\
Scenario 10 & 638 & 392 & 9  & 96  & 0.9075 \\
Scenario 11 & 637 & 378 & 23 & 97  & 0.8942 \\
\bottomrule
\end{tabular}
\end{table}

The confusion matrix components for each scenario are provided in Table \ref{tab:Confusion_Matrix_vit}, including the number of true positives (TP), true negatives (TN), false positives (FP), and false negatives (FN). These values corroborate the overall test accuracy and provide insights into class-wise prediction reliability. Scenario 8 again stands out with a balanced count of TP and TN and a lower FN, contributing to its highest accuracy.
\begin{figure}[htbp]
    \centering

    \begin{subfigure}[b]{0.24\linewidth}
        \centering
        \includegraphics[width=\linewidth]{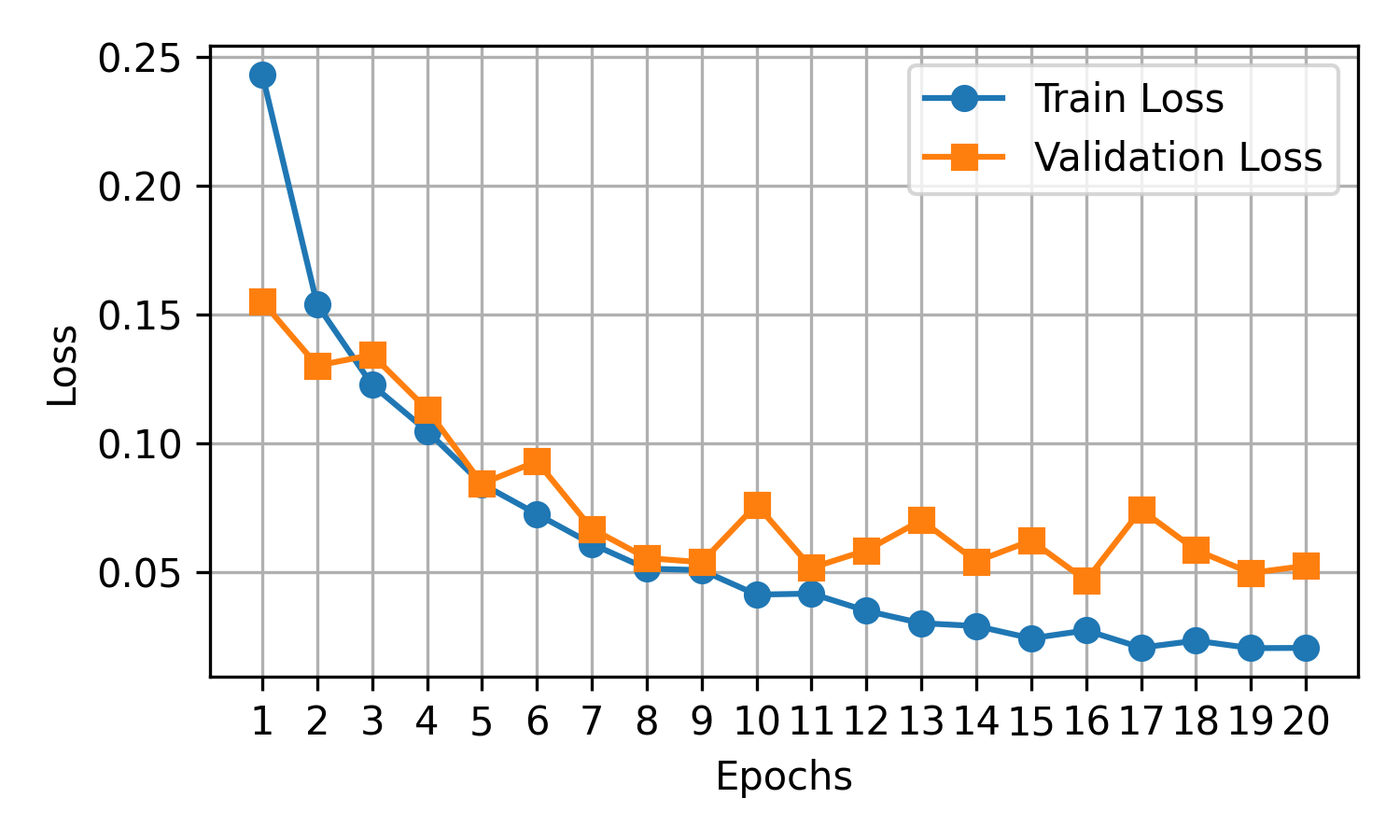}
        \caption{}
        \label{fig:s1}
    \end{subfigure}
    \begin{subfigure}[b]{0.24\linewidth}
        \centering
        \includegraphics[width=\linewidth]{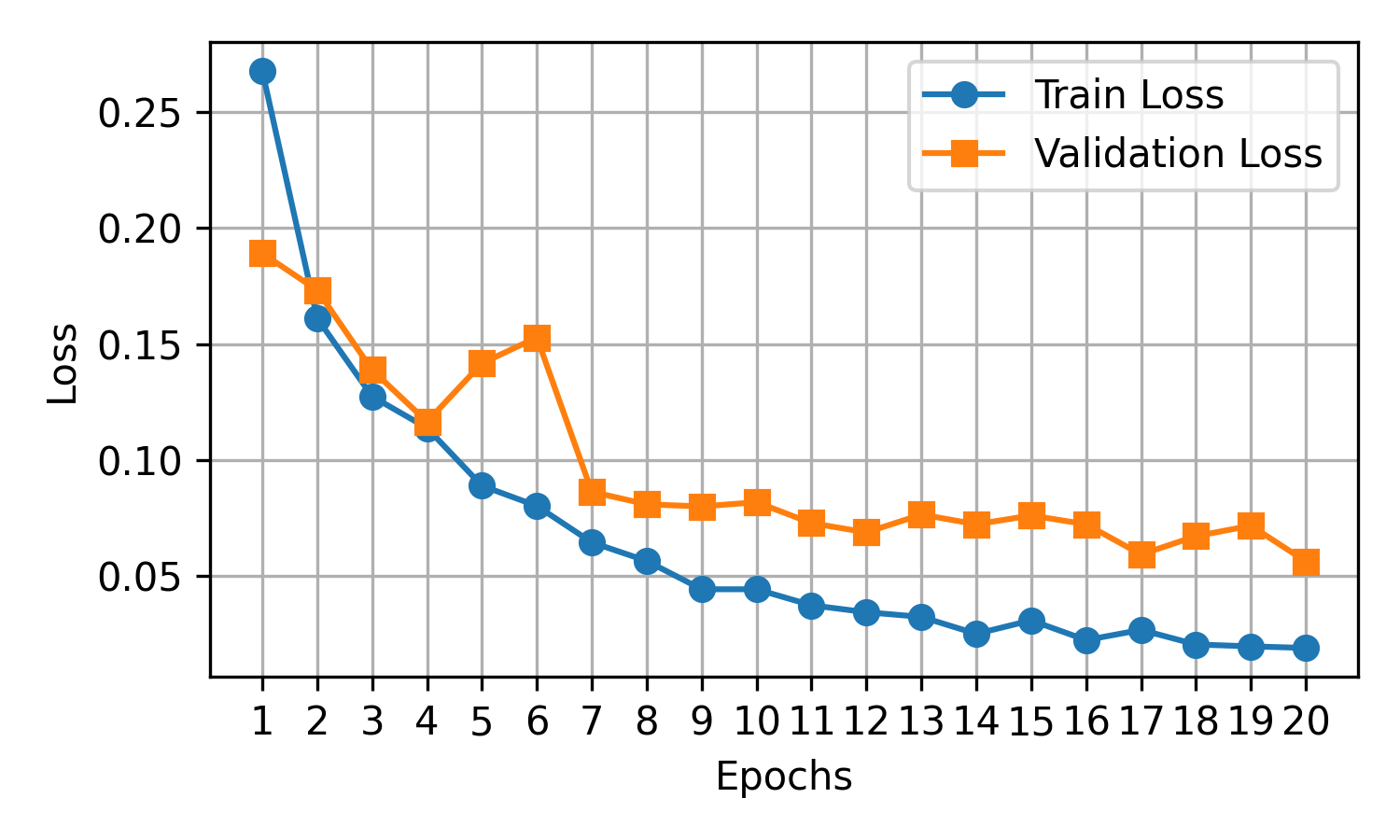}
        \caption{}
        \label{fig:s2}
    \end{subfigure}
    \begin{subfigure}[b]{0.24\linewidth}
        \centering
        \includegraphics[width=\linewidth]{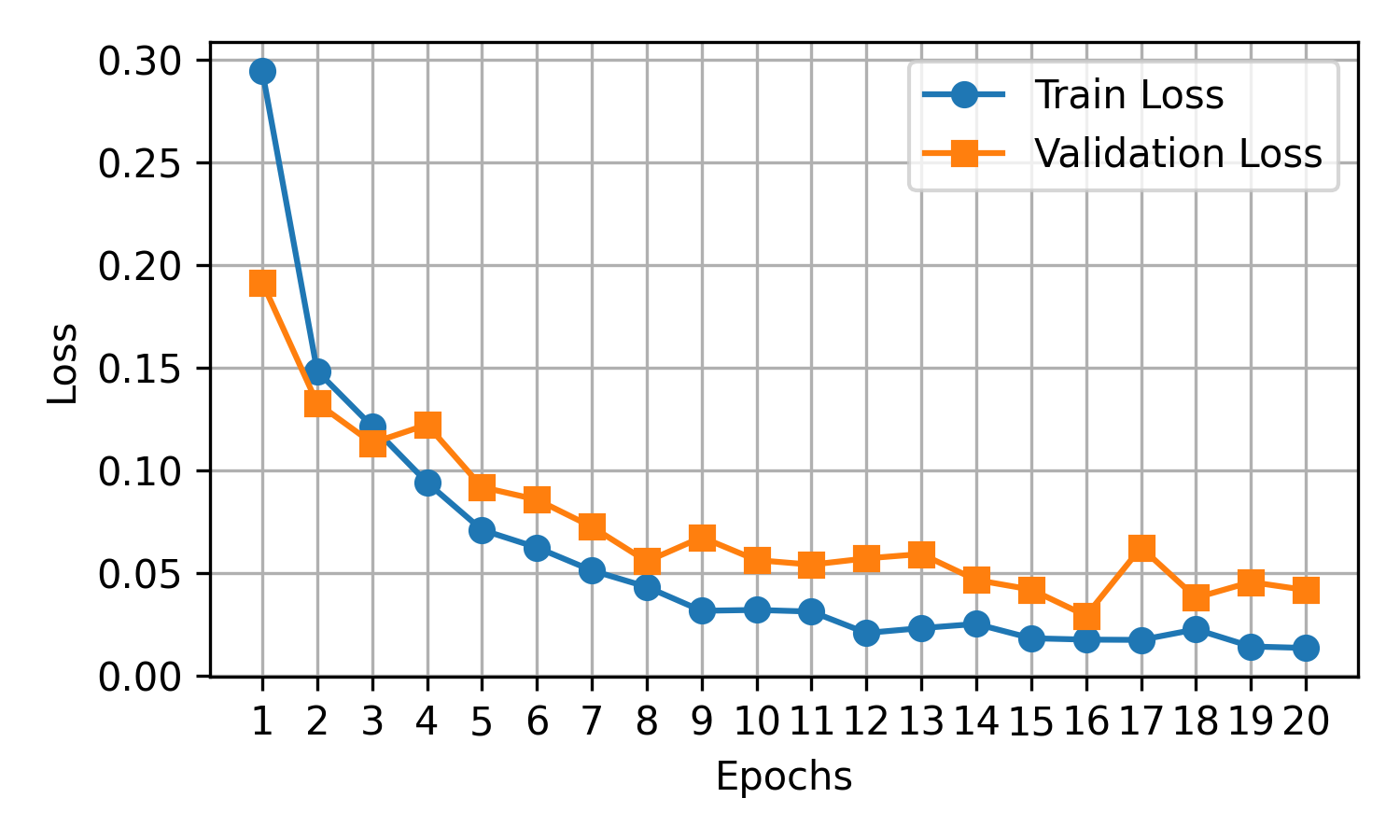}
        \caption{}
        \label{fig:s3}
    \end{subfigure}
    \begin{subfigure}[b]{0.24\linewidth}
        \centering
        \includegraphics[width=\linewidth]{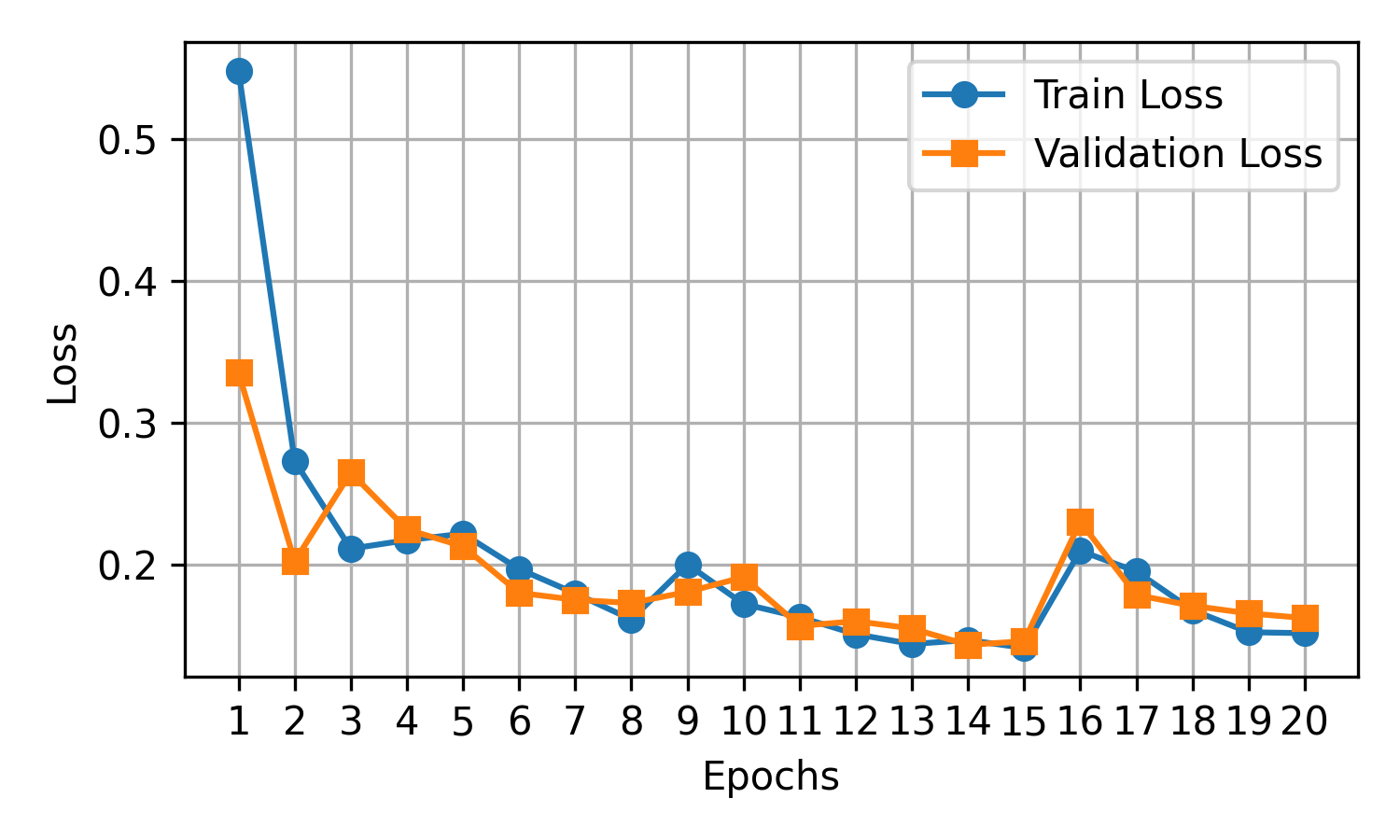}
        \caption{}
        \label{fig:s4}
    \end{subfigure}

    \vspace{0.7em}

    \begin{subfigure}[b]{0.24\linewidth}
        \centering
        \includegraphics[width=\linewidth]{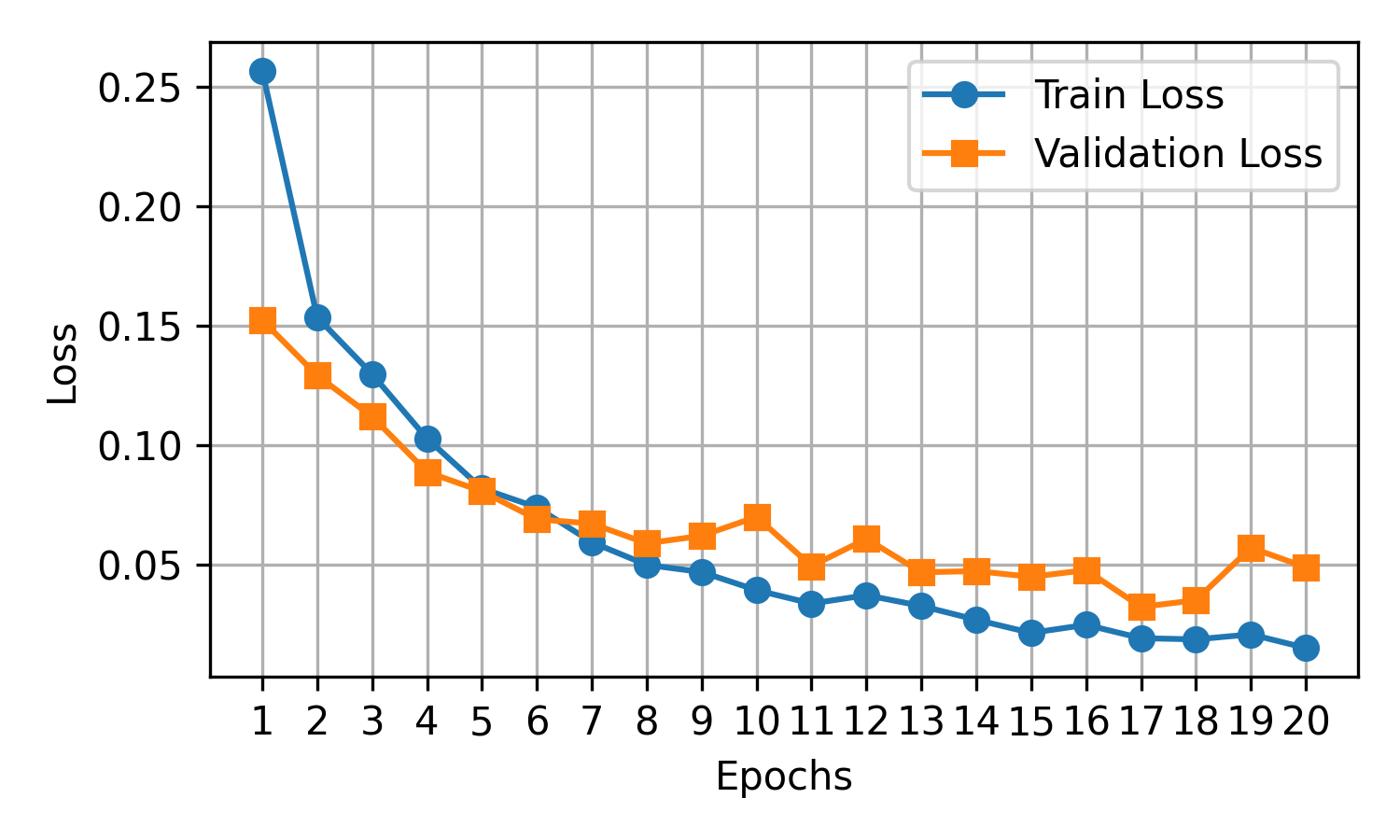}
        \caption{}
        \label{fig:s5}
    \end{subfigure}
    \begin{subfigure}[b]{0.24\linewidth}
        \centering
        \includegraphics[width=\linewidth]{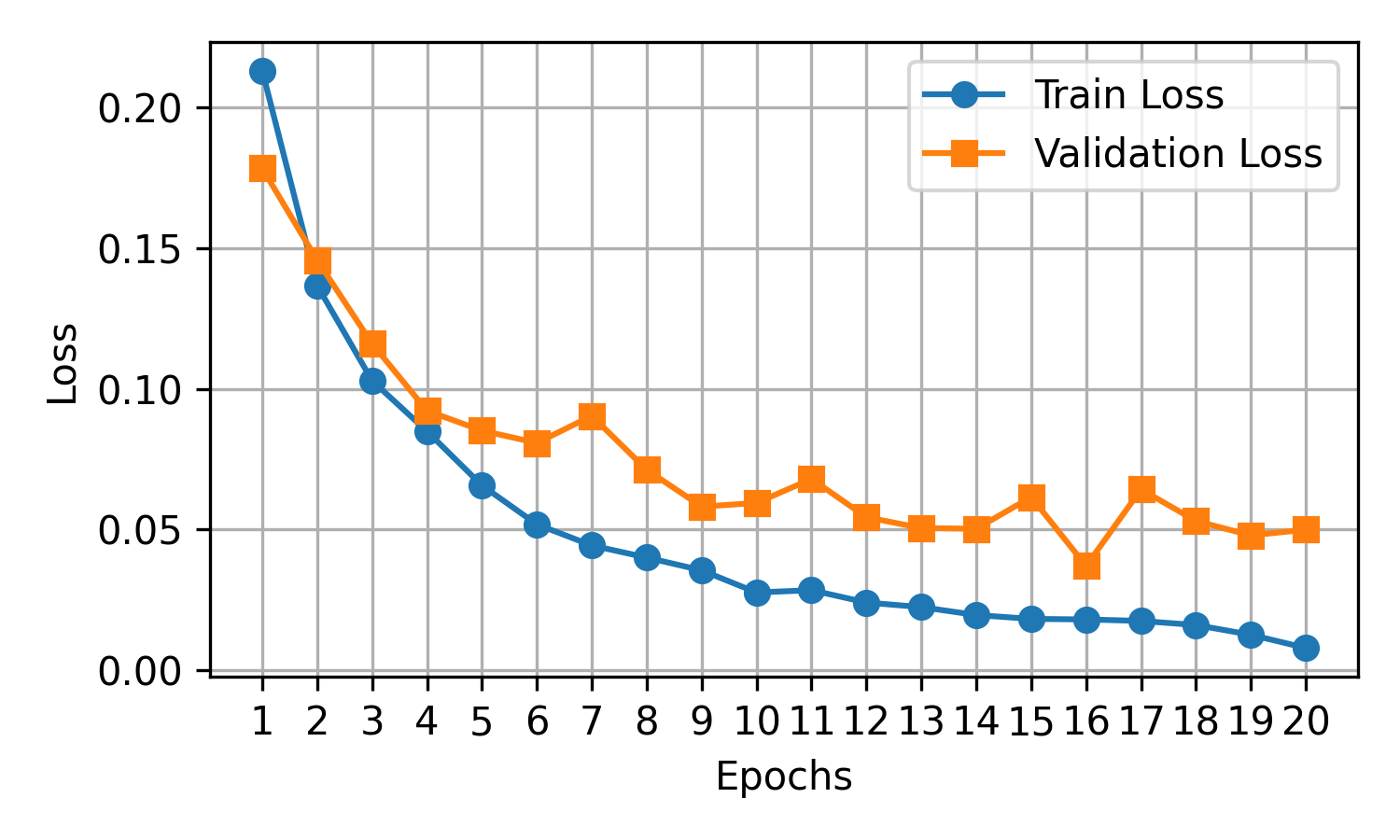}
        \caption{}
        \label{fig:s6}
    \end{subfigure}
    \begin{subfigure}[b]{0.24\linewidth}
        \centering
        \includegraphics[width=\linewidth]{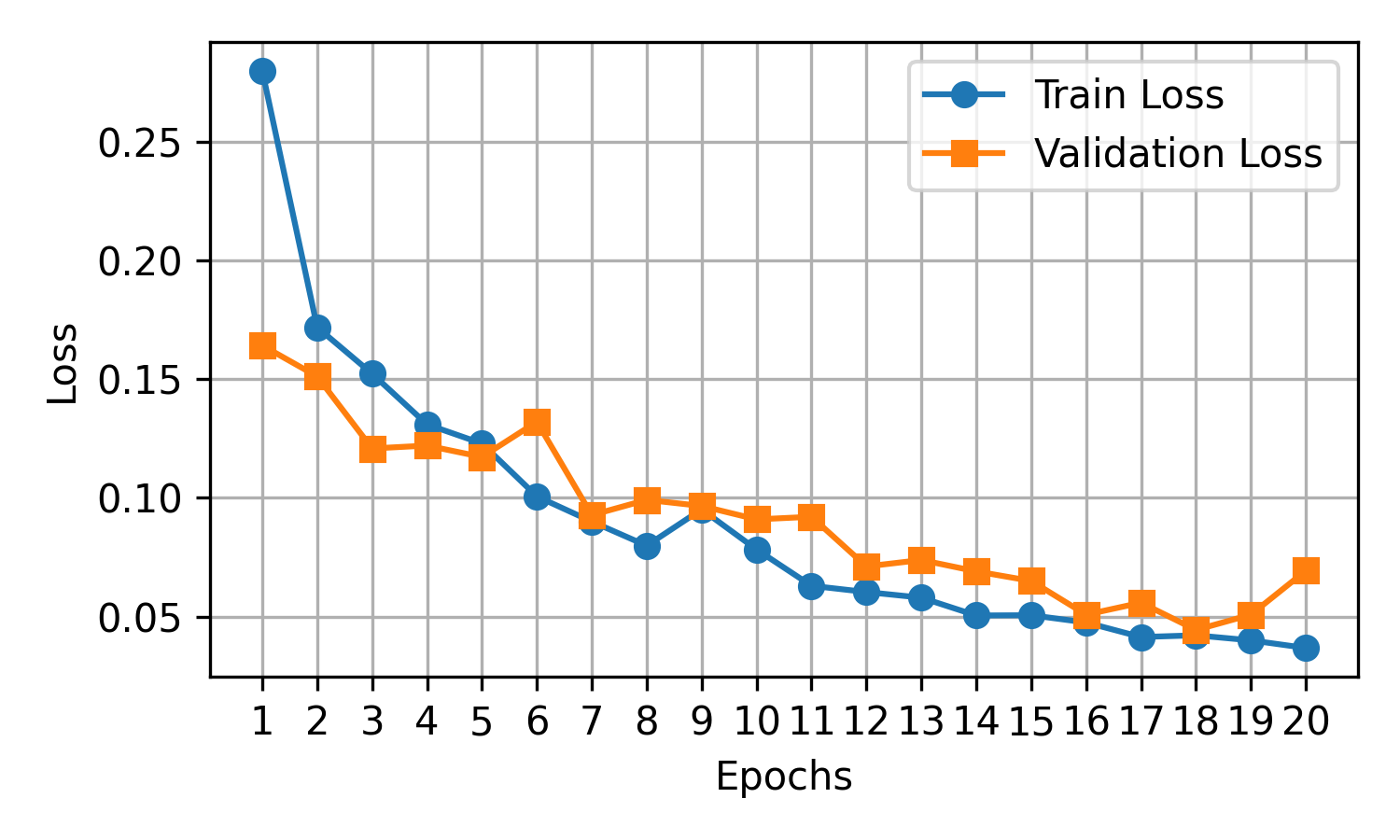}
        \caption{}
        \label{fig:s7}
    \end{subfigure}
    \begin{subfigure}[b]{0.24\linewidth}
        \centering
        \includegraphics[width=\linewidth]{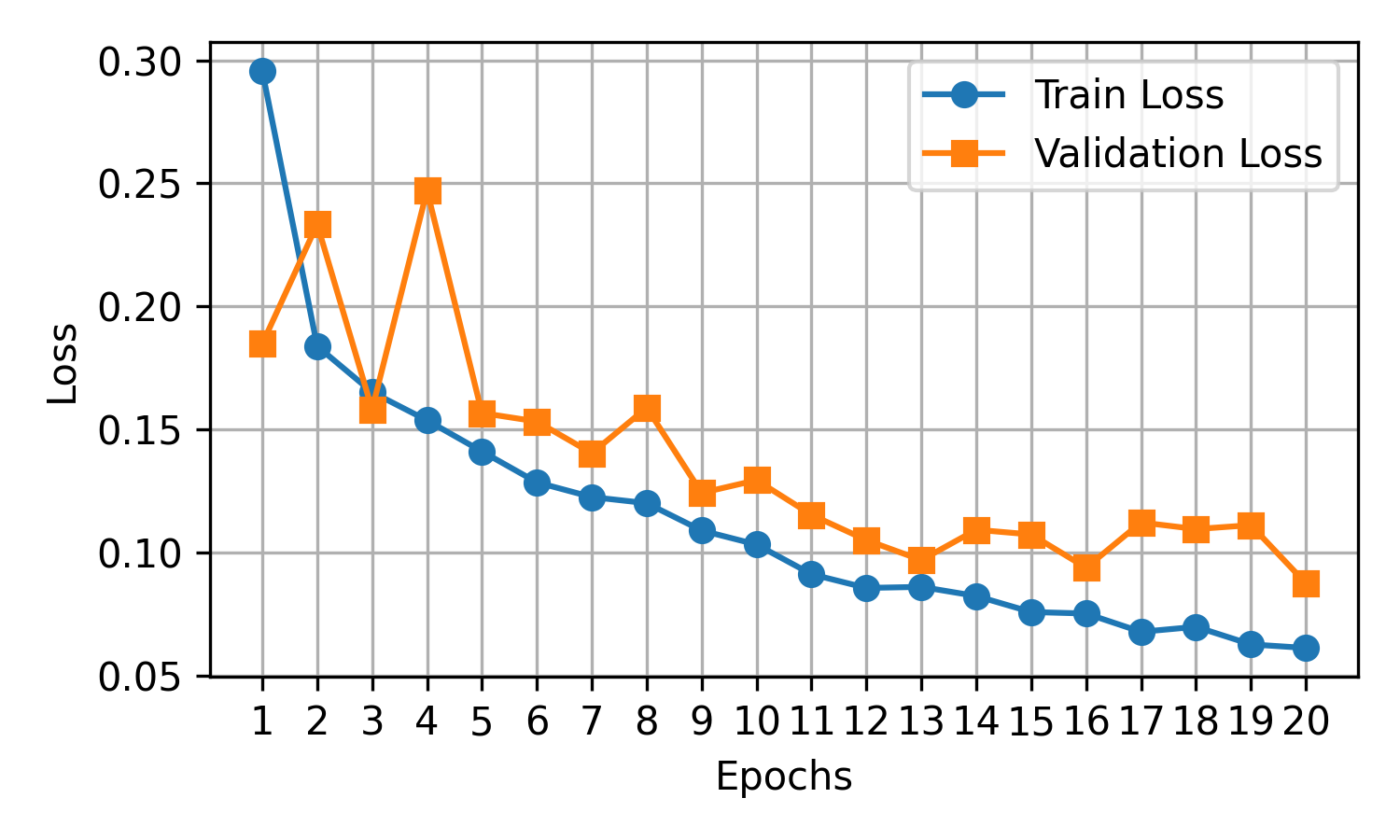}
        \caption{}
        \label{fig:s8}
    \end{subfigure}

    \vspace{0.7em}

    \begin{subfigure}[b]{0.24\linewidth}
        \centering
        \includegraphics[width=\linewidth]{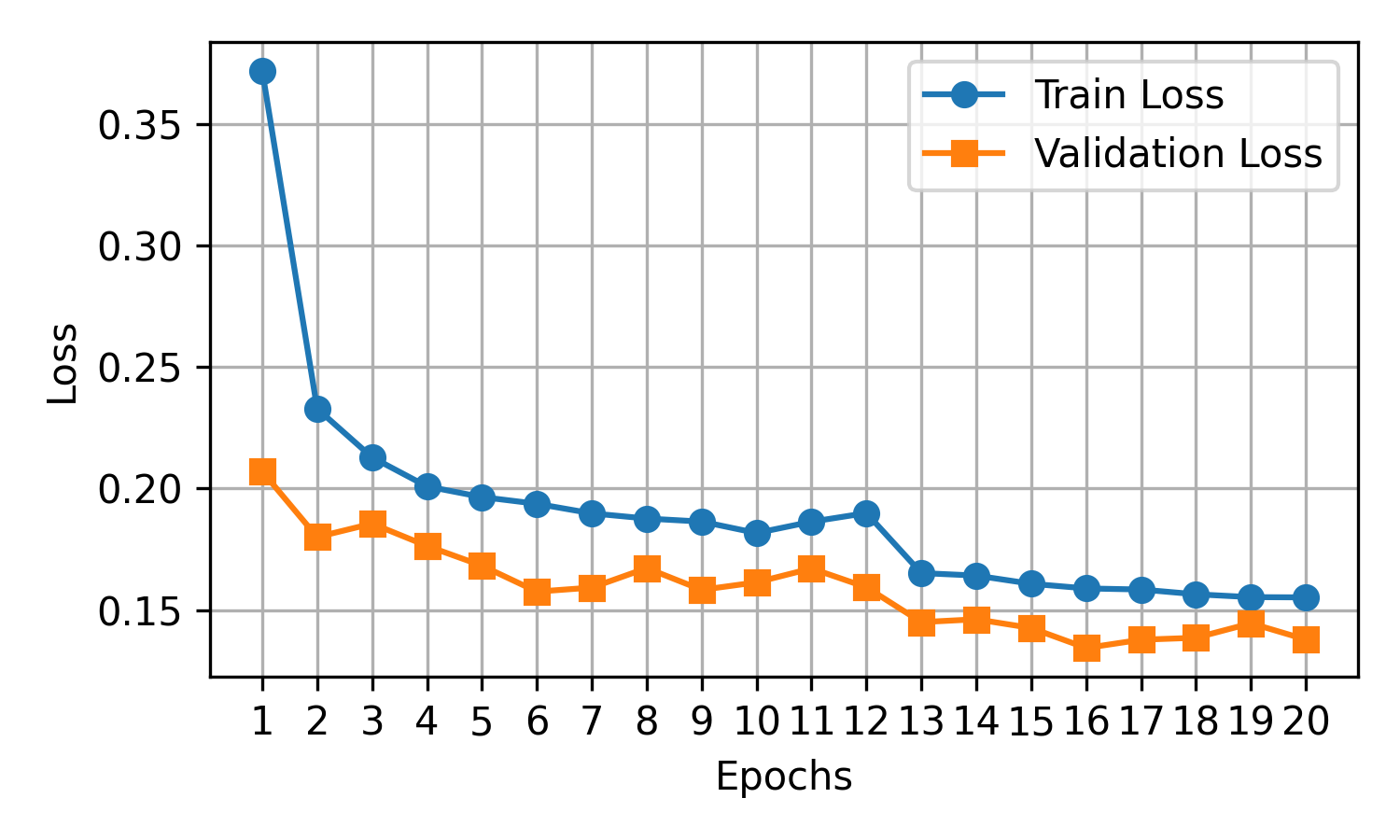}
        \caption{}
        \label{fig:s9}
    \end{subfigure}
    \begin{subfigure}[b]{0.24\linewidth}
        \centering
        \includegraphics[width=\linewidth]{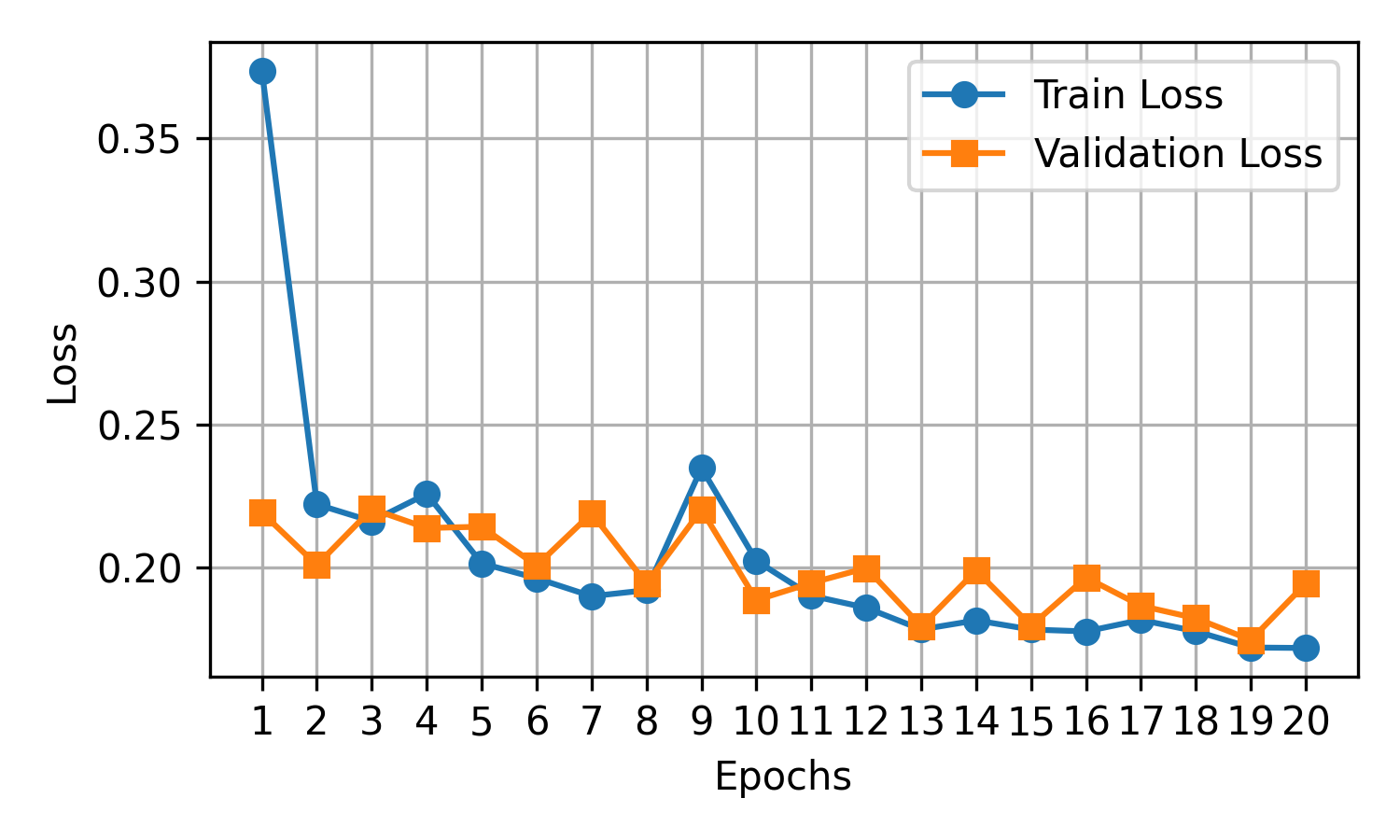}
        \caption{}
        \label{fig:s10}
    \end{subfigure}
    \begin{subfigure}[b]{0.24\linewidth}
        \centering
        \includegraphics[width=\linewidth]{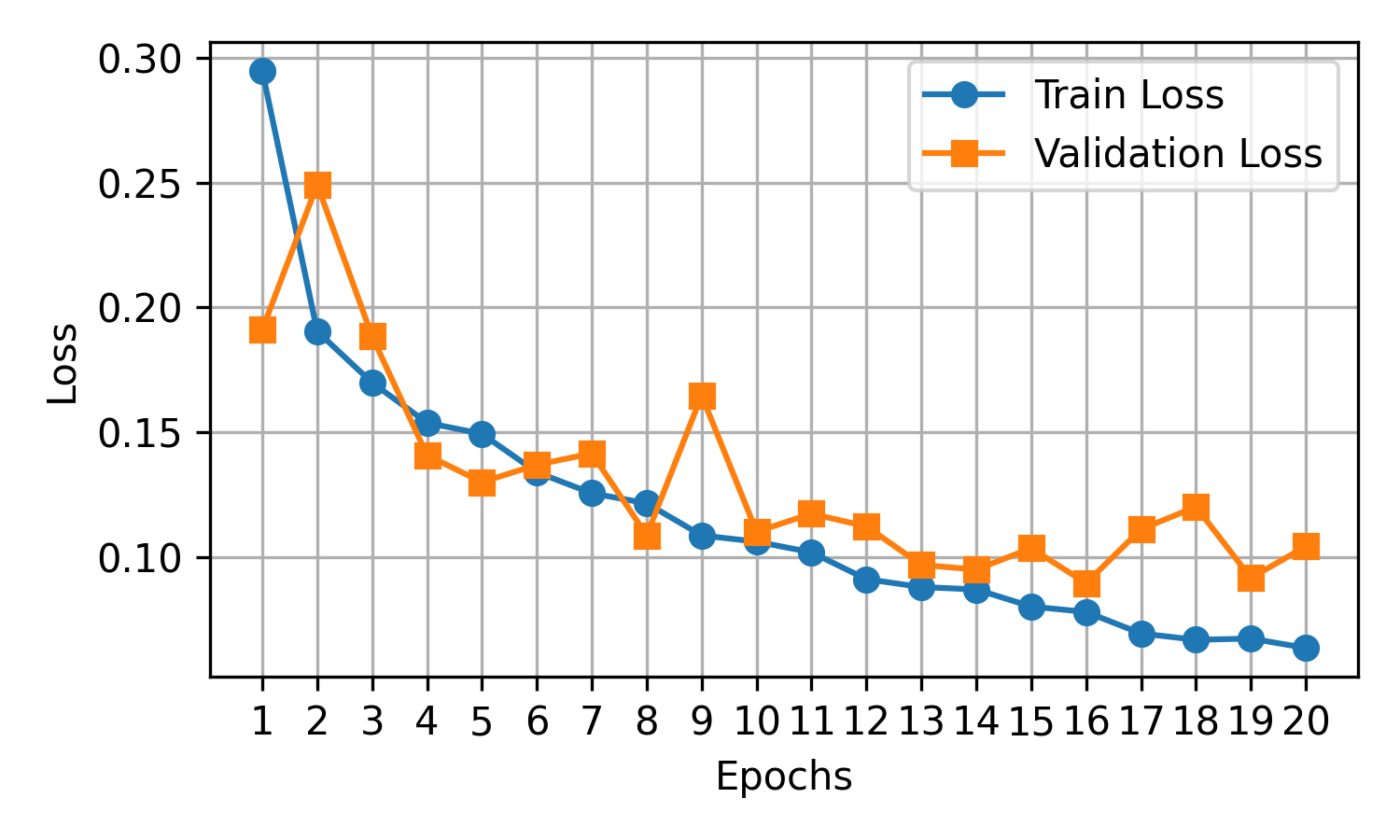}
        \caption{}
        \label{fig:s11}
    \end{subfigure}
    
    \caption{Loss curves for 11 scenarios: Fig. a–k corresponding to scenario 1 to 11.}
    \label{fig:loss_samples}
\end{figure}

\begin{figure}[hbt!]
    \centering

    \begin{subfigure}[b]{0.24\linewidth}
        \centering
        \includegraphics[width=\linewidth]{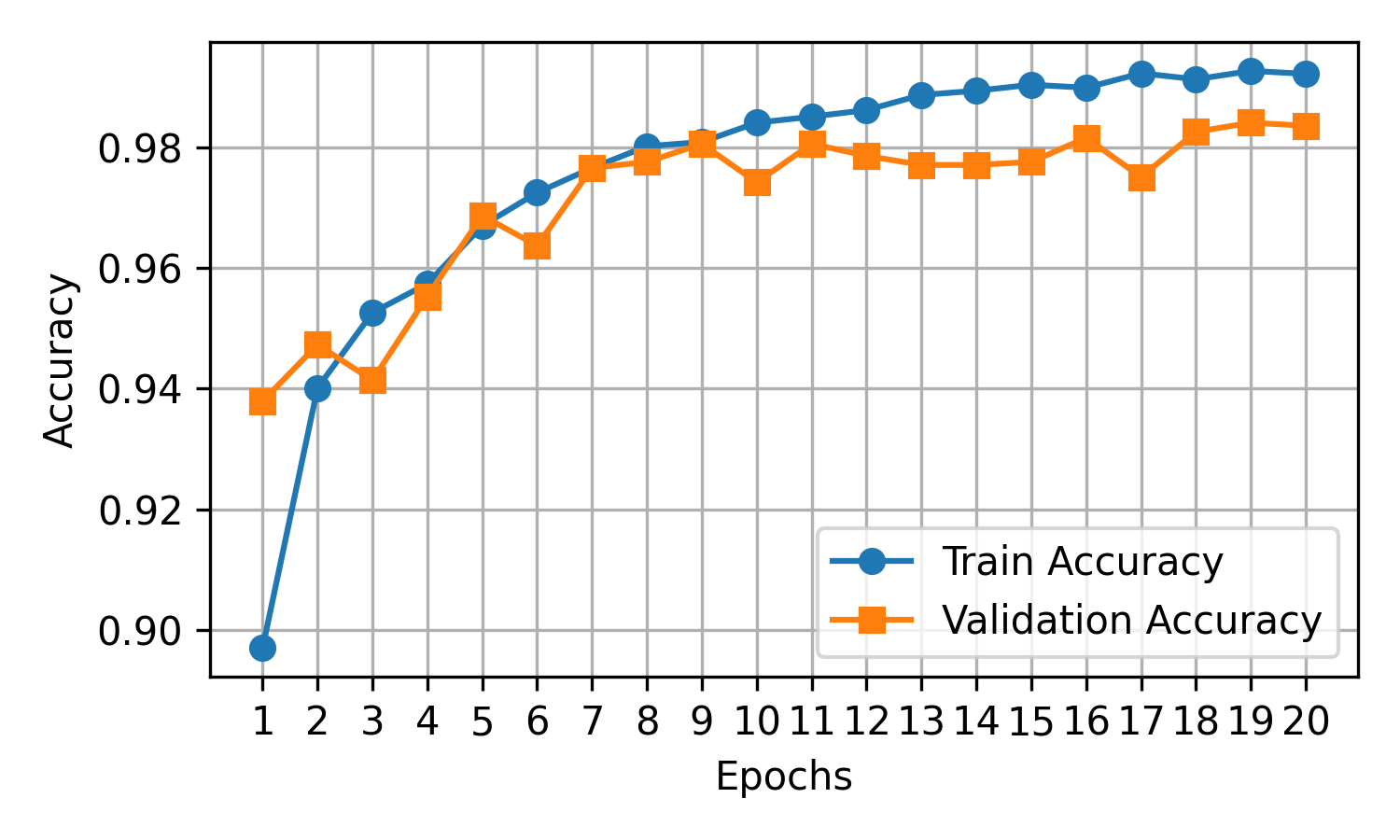}
        \caption{}
        \label{fig:s1}
    \end{subfigure}
    \begin{subfigure}[b]{0.24\linewidth}
        \centering
        \includegraphics[width=\linewidth]{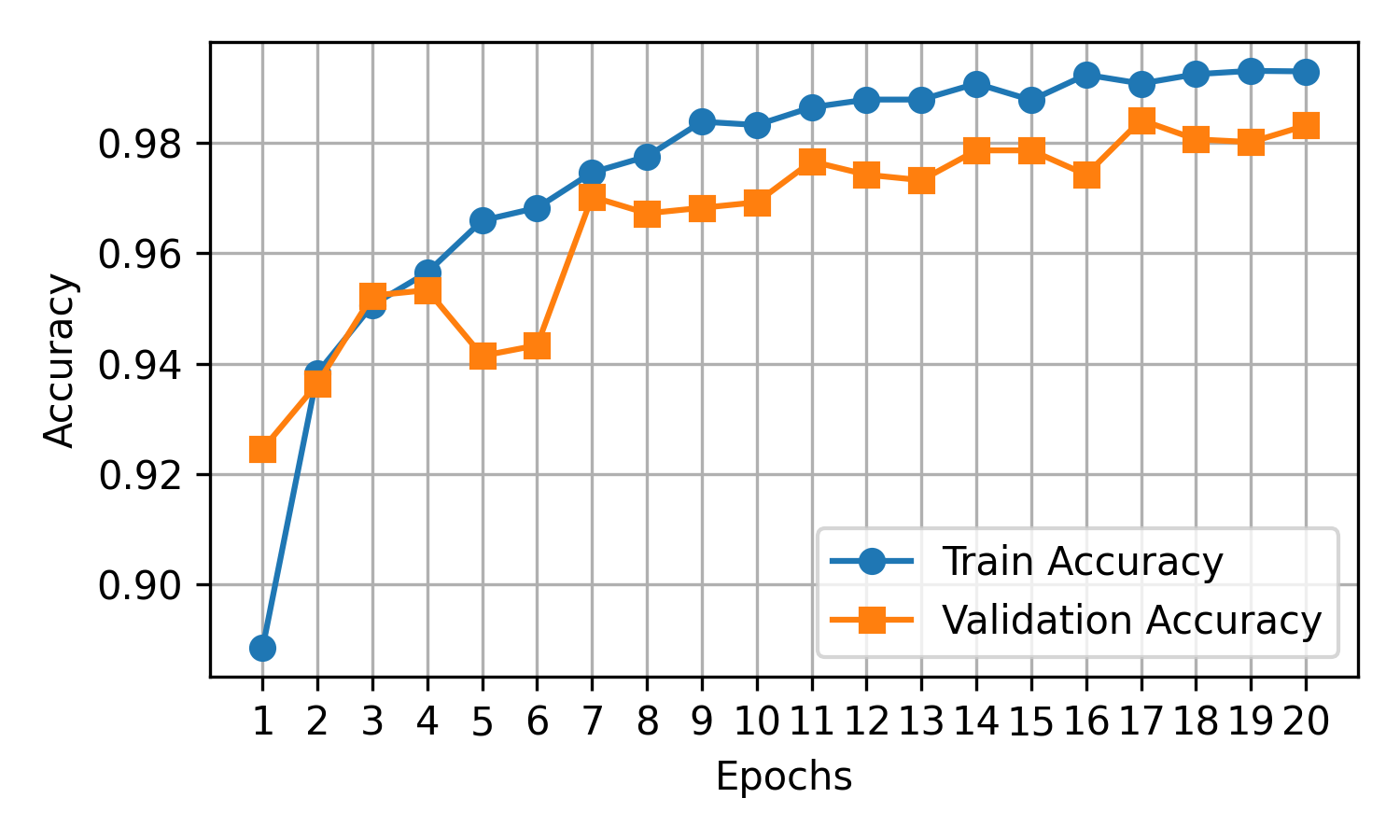}
        \caption{}
        \label{fig:s2}
    \end{subfigure}
    \begin{subfigure}[b]{0.24\linewidth}
        \centering
        \includegraphics[width=\linewidth]{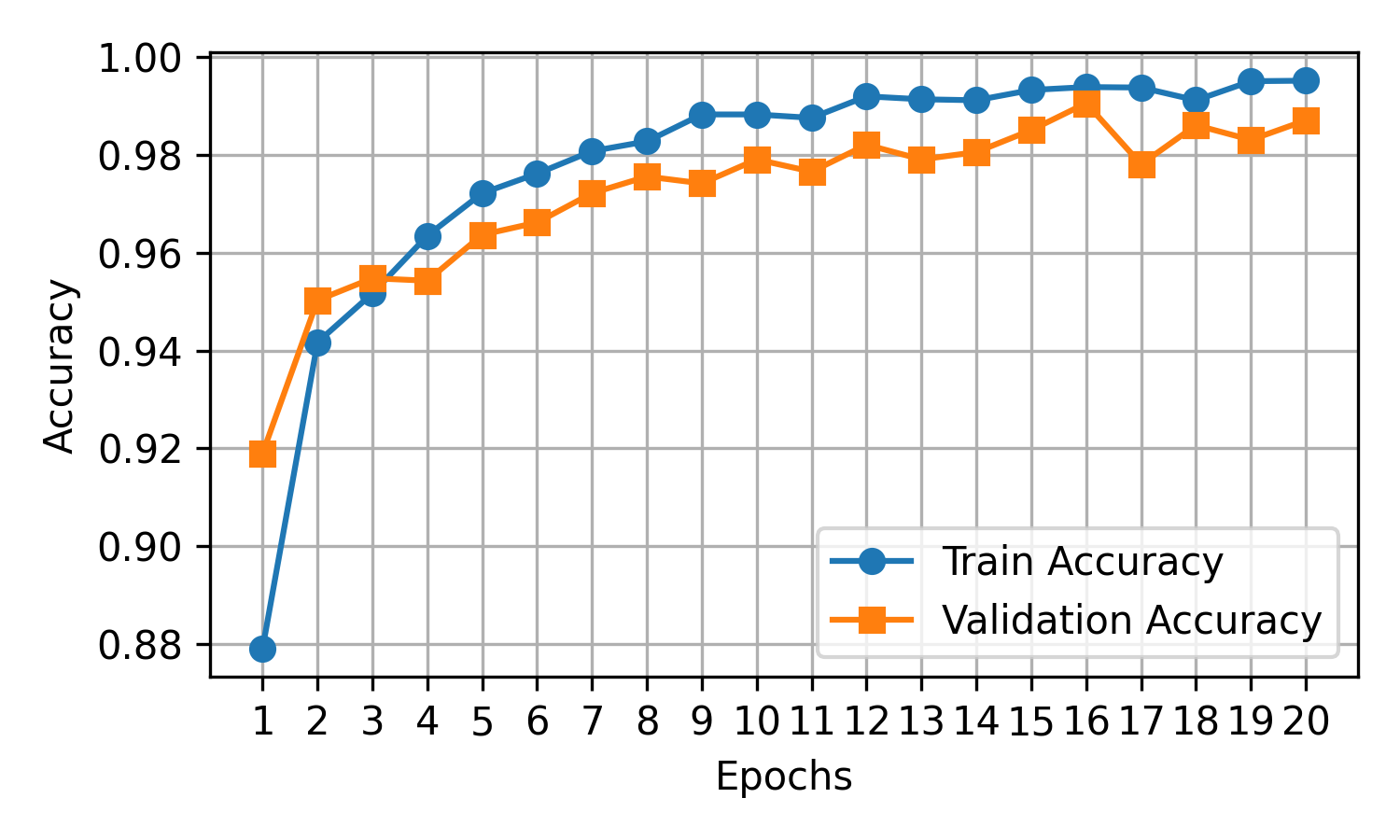}
        \caption{}
        \label{fig:s3}
    \end{subfigure}
    \begin{subfigure}[b]{0.24\linewidth}
        \centering
        \includegraphics[width=\linewidth]{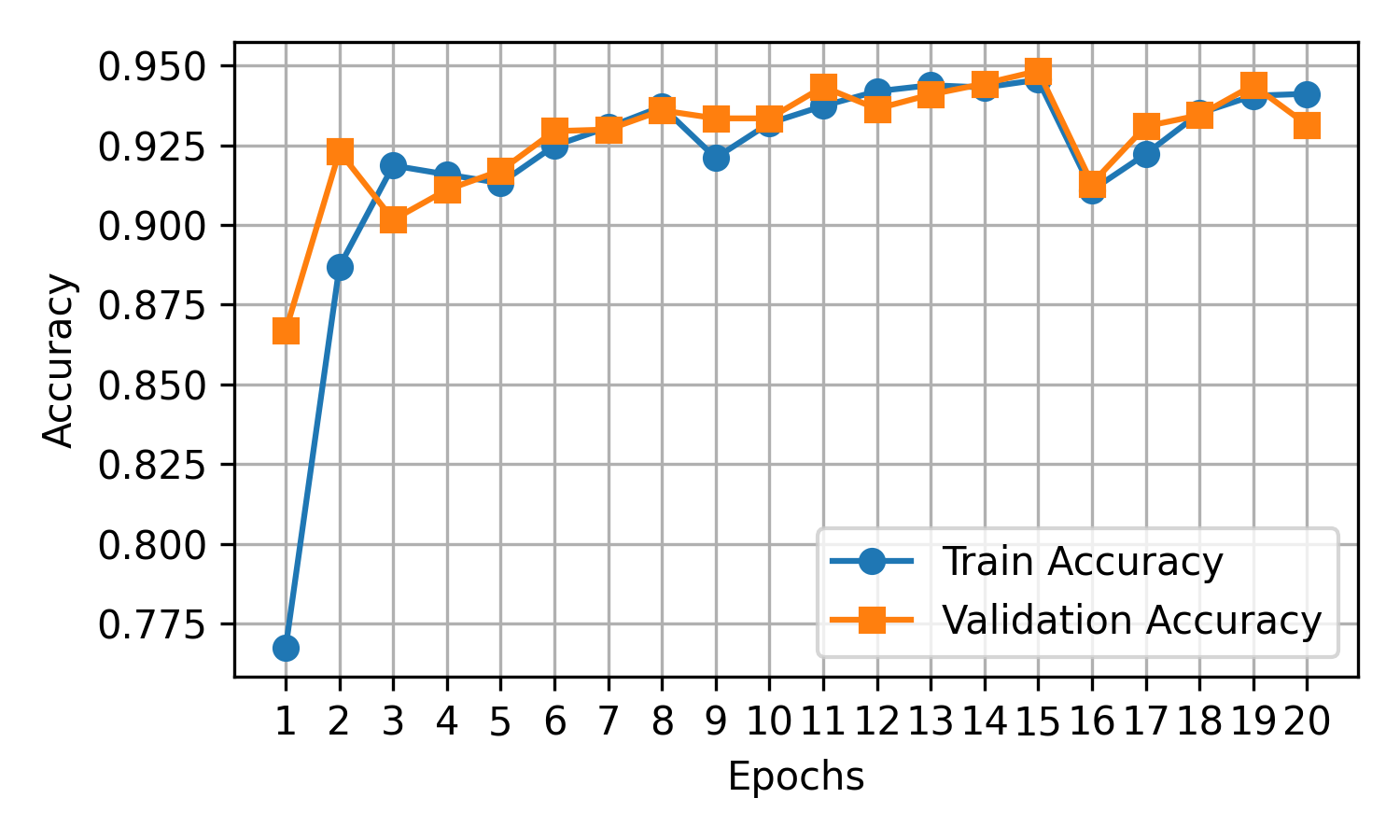}
        \caption{}
        \label{fig:s4}
    \end{subfigure}

    \vspace{0.7em}

    \begin{subfigure}[b]{0.24\linewidth}
        \centering
        \includegraphics[width=\linewidth]{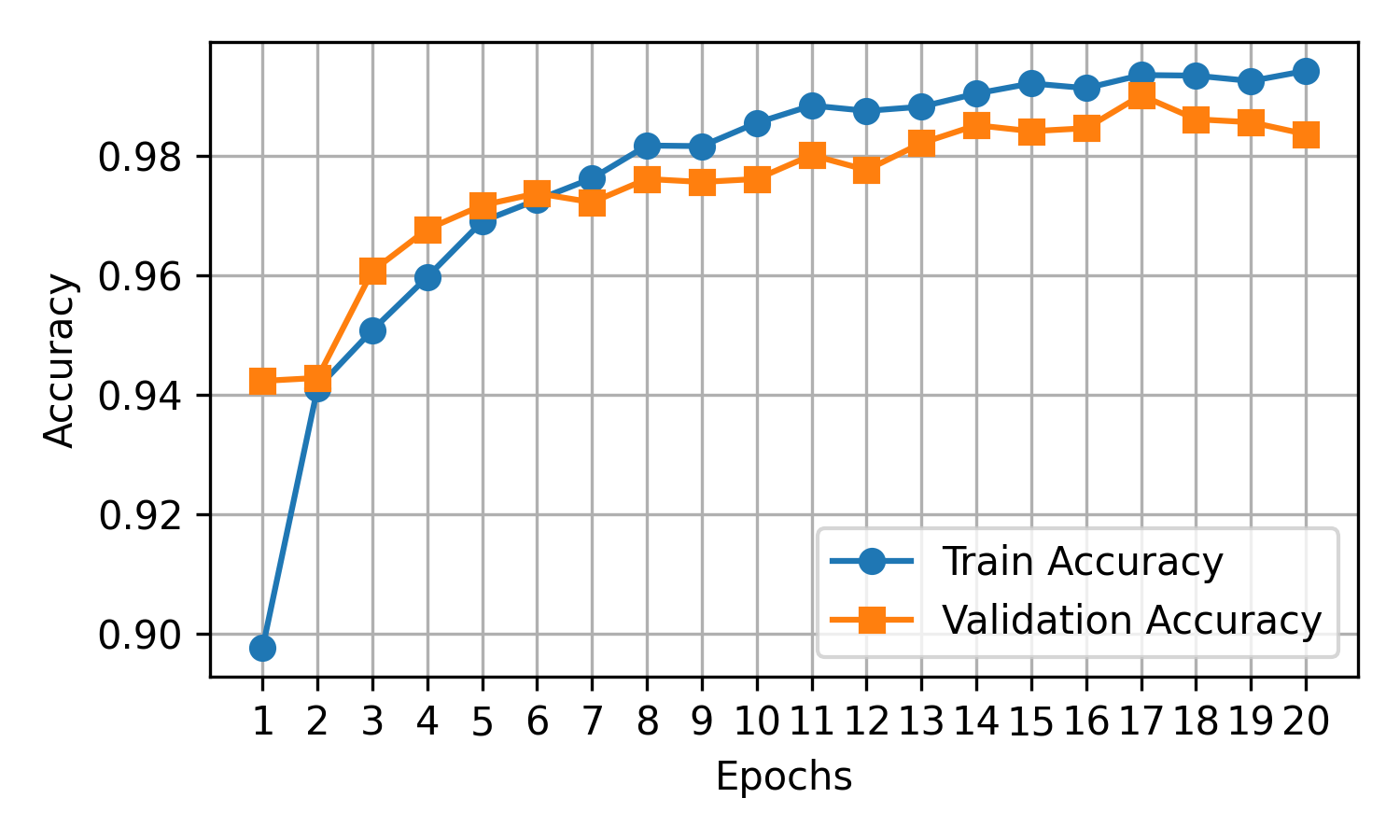}
        \caption{}
        \label{fig:s5}
    \end{subfigure}
    \begin{subfigure}[b]{0.24\linewidth}
        \centering
        \includegraphics[width=\linewidth]{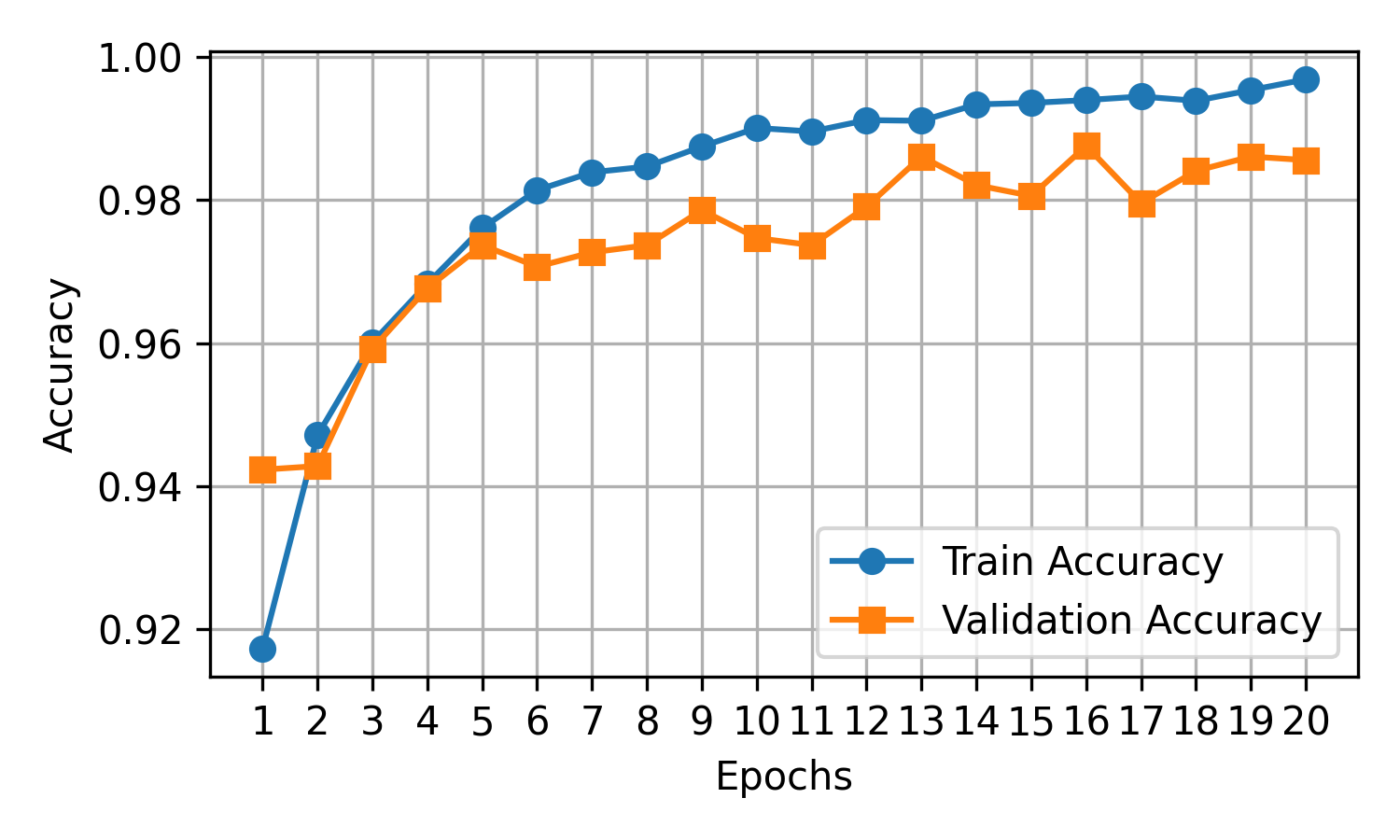}
        \caption{}
        \label{fig:s6}
    \end{subfigure}
    \begin{subfigure}[b]{0.24\linewidth}
        \centering
        \includegraphics[width=\linewidth]{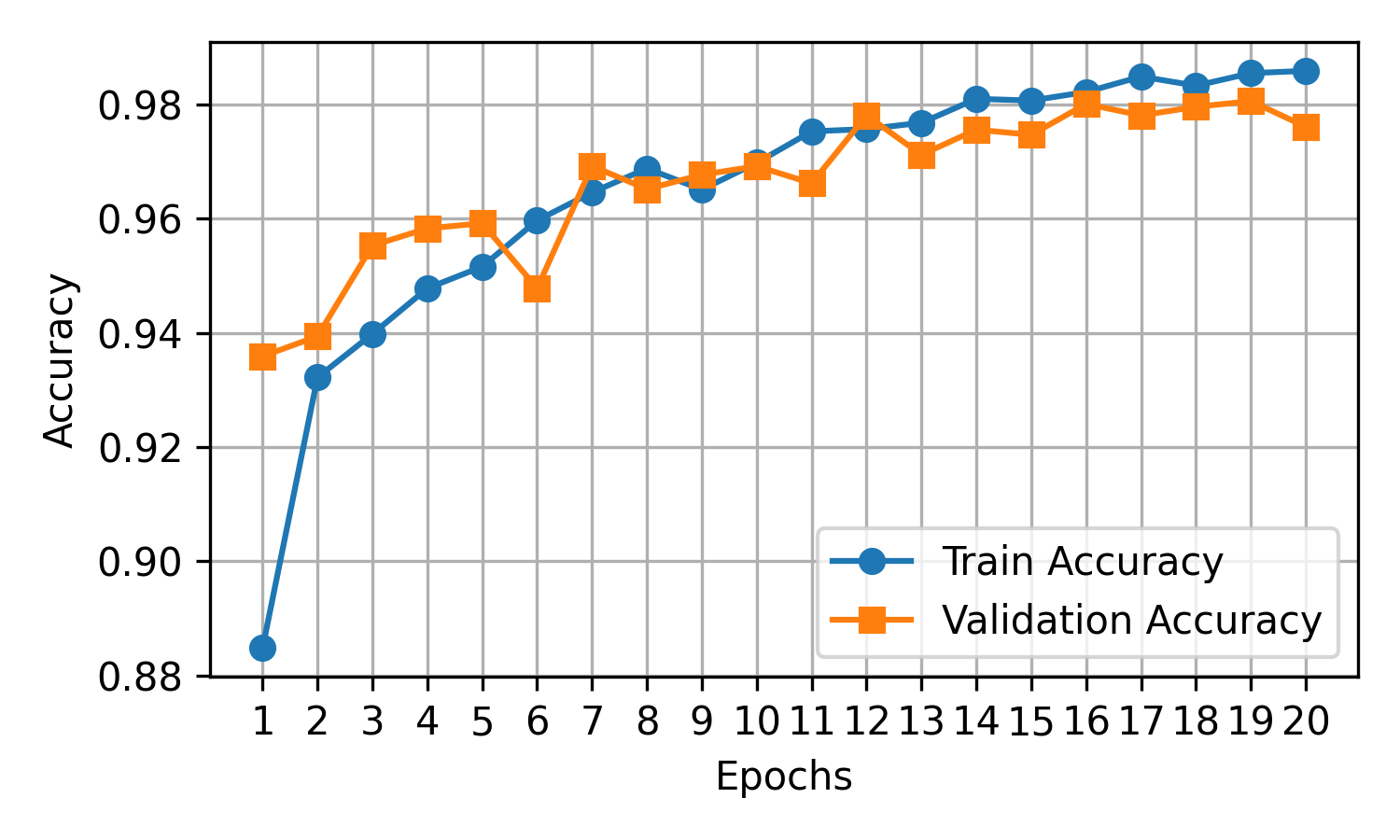}
        \caption{}
        \label{fig:s7}
    \end{subfigure}
    \begin{subfigure}[b]{0.24\linewidth}
        \centering
        \includegraphics[width=\linewidth]{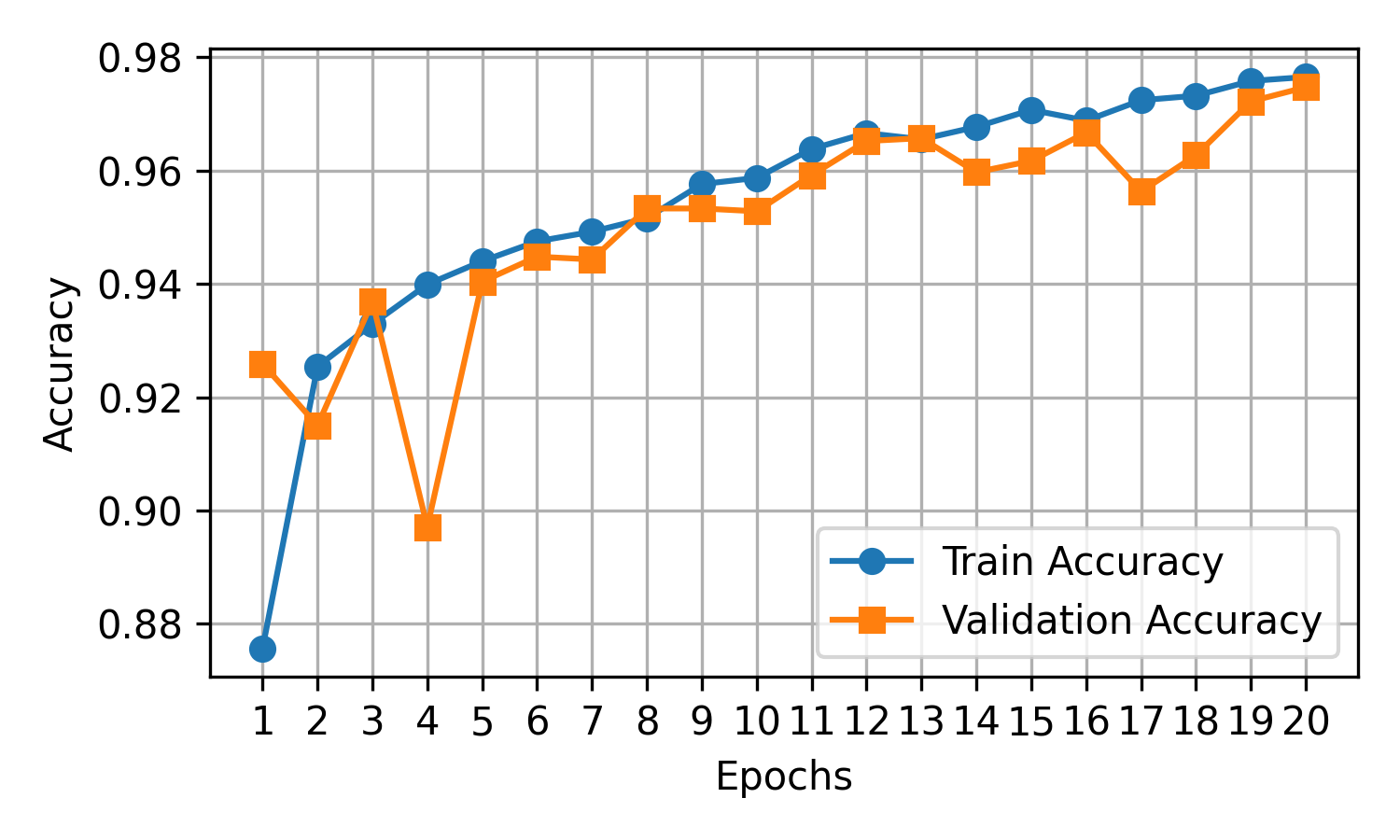}
        \caption{}
        \label{fig:s8}
    \end{subfigure}

    \vspace{0.7em}

    \begin{subfigure}[b]{0.24\linewidth}
        \centering
        \includegraphics[width=\linewidth]{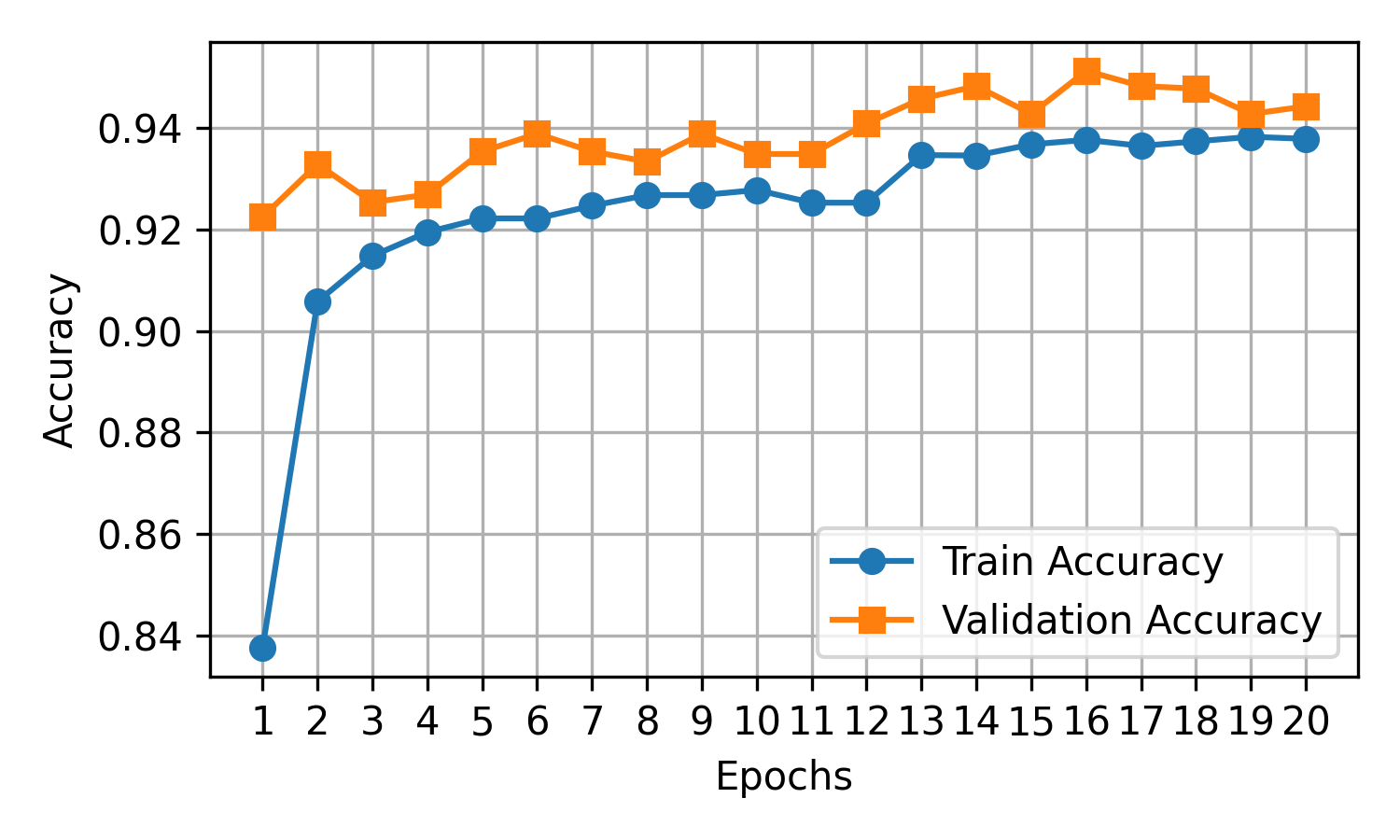}
        \caption{}
        \label{fig:s9}
    \end{subfigure}
    \begin{subfigure}[b]{0.24\linewidth}
        \centering
        \includegraphics[width=\linewidth]{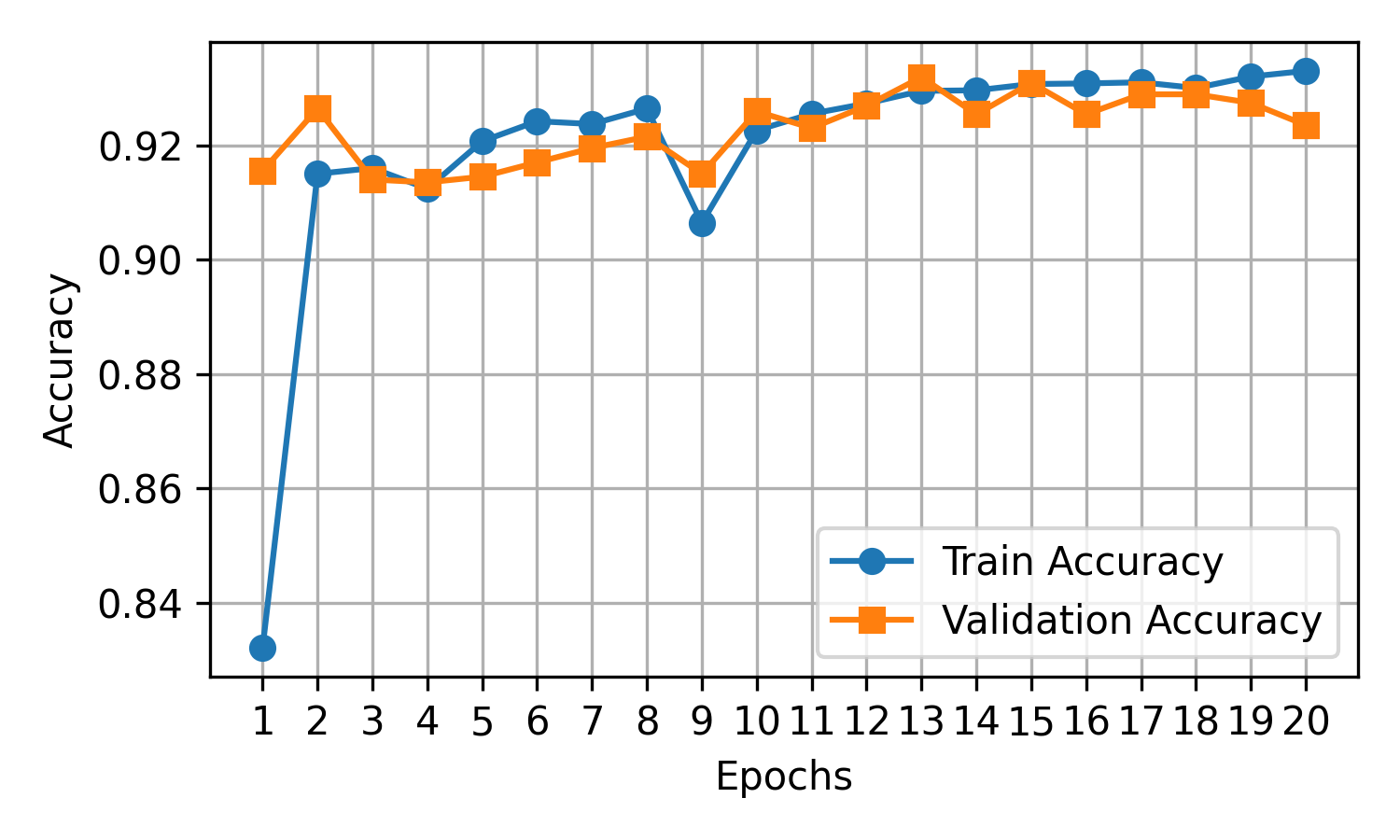}
        \caption{}
        \label{fig:s10}
    \end{subfigure}
    \begin{subfigure}[b]{0.24\linewidth}
        \centering
        \includegraphics[width=\linewidth]{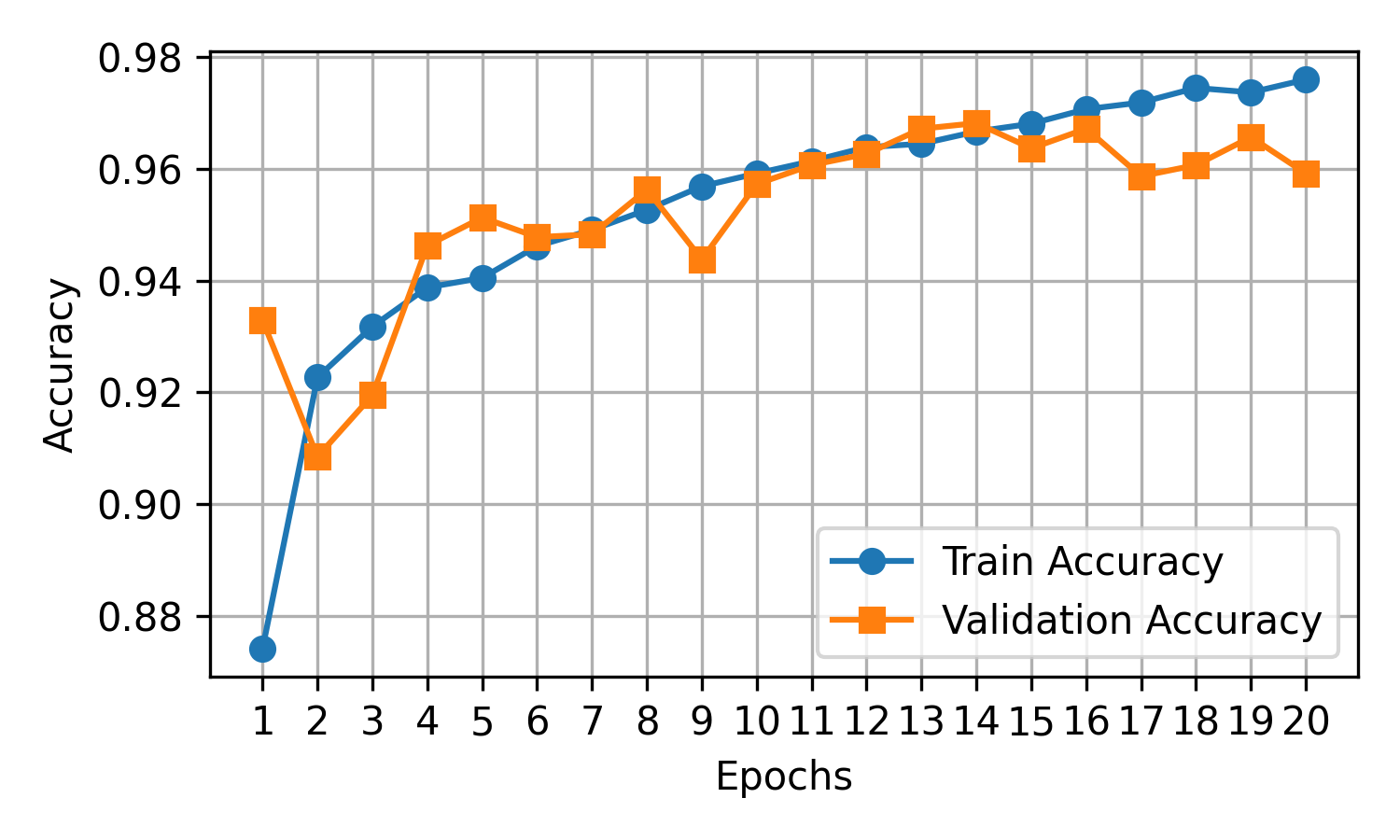}
        \caption{}
        \label{fig:s11}
    \end{subfigure}
    
    \caption{Accuracy curves for 11 scenarios: Fig. a–k corresponding to scenario 1 to 11.}
    \label{fig:accuracy_samples}
\end{figure}

The loss and accuracy curves for all 11 scenarios are depicted in Fig. \ref{fig:loss_samples} and Fig. \ref{fig:accuracy_samples}, respectively. These plots offer visual confirmation of training convergence and generalization. From the loss and accuracy curves, it is evident that most scenarios converge smoothly, with minimal overfitting. Scenario 5 is a notable exception, exhibiting a pronounced gap between training and validation performance. This is further reflected in its confusion matrix, where the number of false negatives (117) significantly exceeds other scenarios, explaining its poor generalization (test accuracy: 88.19\%). Scenario 8, despite modest fluctuations during training, achieves the highest test accuracy (91.62\%) and maintains a balanced distribution of true and false predictions. Scenarios 9 and 10 notably report the lowest false positive counts (11 and 9, respectively), making them suitable for applications where false alarms are critical. In contrast, Scenario 7 shows the highest recall (TP: 663, FN: 71), which is necessary in settings where missing stressed cases could be detrimental.

This systematic experimentation helped us analyze the trade-off between model complexity and performance. The results showed that fine-tuning only the last 2–3 encoder blocks of ViT-B/16, combined with suitable regularization, yielded performance comparable to that of training the entire network or using ViT-L/16, but with significantly fewer trainable parameters and faster training times.

\subsubsection{Analyzing Attention Maps}
Visualizing attention weights provides insights into how the ViT focus on different parts of an input during processing. By examining these weights, researchers and practitioners can understand the model's decision-making process, diagnose potential biases, and improve interpretability.

The following pseudocode outlines a systematic approach to calculate and visualize attention weights from a Vision Transformer model. This process involves capturing attention weights during the forward pass, computing attention scores, and generating visual representations of these scores.

\textbf{Initialization:}
The process begins by initializing an empty list to store the attention weights that will be captured during the forward pass of the Vision Transformer model. This list will later be used to compute and visualize the attention maps.

\textbf{Forward Pass and Capture Attention Weights:}
The next step involves iterating through each layer of the Vision Transformer model. For each layer, a hook is registered to capture the attention weights. The input image is then passed through the Vision Transformer to compute the output features. This stage ensures that attention weights are collected during the forward pass for later analysis.

\textbf{Calculate Attention Score:}
After capturing the attention weights, the algorithm processes each weight to compute the attention scores. This involves extracting the query, key, and value tensors from the hook outputs. The attention score is computed as per the principles discussed in section \ref{sec:atten}. The attention scores are then normalized to ensure they are in a range suitable for visualization.

\textbf{Visualize Attention Maps:}
In the visualization phase, each normalized attention map is resized to match the dimensions of the input image. A \textit{colormap} (e.g., 'hot') is applied to the attention map to highlight areas of high attention. The attention map is then overlaid on the original image to create a visual representation of where the model is focusing.

\textbf{Display:}
Finally, both the original image and the overlaid attention maps are displayed, allowing for an interpretation of how the Vision Transformer model is making its decisions based on different regions of the input image.

\begin{algorithm}[H]
\caption{Calculate and Visualize Attention Weights}
\label{alg:attention_visualization}

\KwIn{Input image, Vision Transformer model}
\KwOut{Attention maps visualization}
\setcounter{AlgoLine}{0}

\SetKwFunction{FMain}{main}
\SetKwProg{Fn}{Function}{:}{}
\Fn{\FMain}{
    \KwData{Input image, Vision Transformer model}
    
    \tcp{Forward Pass and Capture Attention Weights}
    Initialize attention weights list \\
    \For{each layer in Vision Transformer}{
        Register hook to capture attention weights \\
        Pass input image through Vision Transformer \\
        Compute output features
    }
    
    \tcp{Calculate Attention Score}
    \For{each captured attention weight}{
        Extract query, key, value from the hook output \\
        Compute attention score as \texttt{Attention = Query $\times$ Key\textsuperscript{T}} \\
        \texttt{Attention(Q,K,V) = A $\times$ V} \\
        Normalize attention score
    }
    
    \tcp{Visualize Attention Maps}
    \For{each attention score}{
        Resize attention map to match input image dimensions \\
        Apply colormap (e.g., 'hot') to visualize attention weights \\
        Overlay attention map on original image
    }
    
    Display the original image and attention maps
}

\end{algorithm}

\begin{figure}[htbp]
    \centering
    
    \begin{subfigure}[b]{0.24\linewidth}
        \centering
        \includegraphics[width=\linewidth]{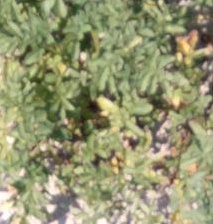}
        \caption{}
        \label{fig:subfig1}
    \end{subfigure}
    
    \vspace{1em}
    \begin{subfigure}[b]{0.24\linewidth}
        \centering
        \includegraphics[width=\linewidth]{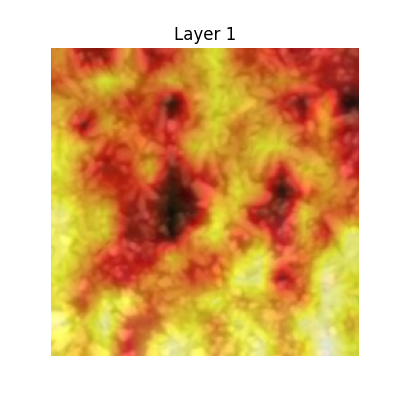}
        \caption{}
        \label{fig:subfig2}
    \end{subfigure}
    \begin{subfigure}[b]{0.24\linewidth}
        \centering
        \includegraphics[width=\linewidth]{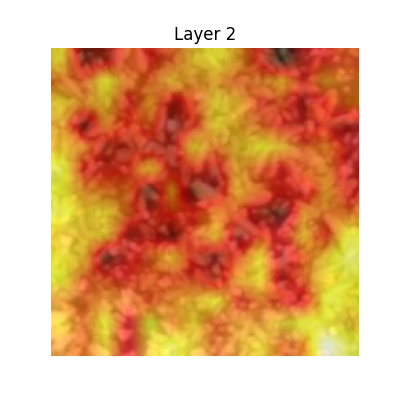}
        \caption{}
        \label{fig:subfig3}
    \end{subfigure}
    \begin{subfigure}[b]{0.24\linewidth}
        \centering
        \includegraphics[width=\linewidth]{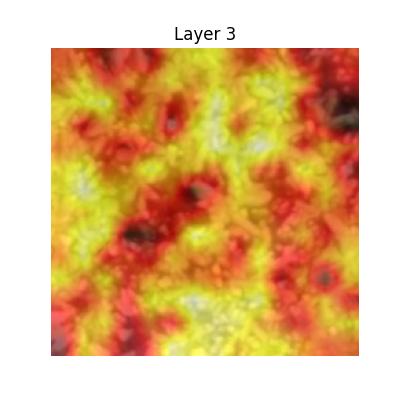}
        \caption{}
        \label{fig:subfig4}
    \end{subfigure}
    \begin{subfigure}[b]{0.24\linewidth}
        \centering
        \includegraphics[width=\linewidth]{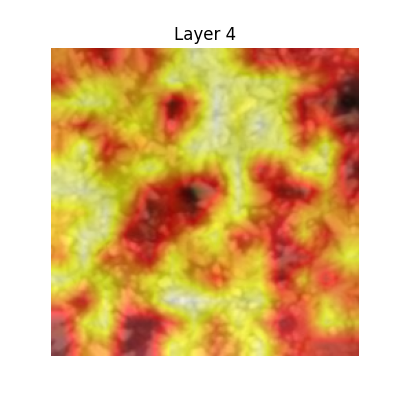}
        \caption{}
        \label{fig:subfig5}
    \end{subfigure}

    \vspace{1em}
    \begin{subfigure}[b]{0.24\linewidth}
        \centering
        \includegraphics[width=\linewidth]{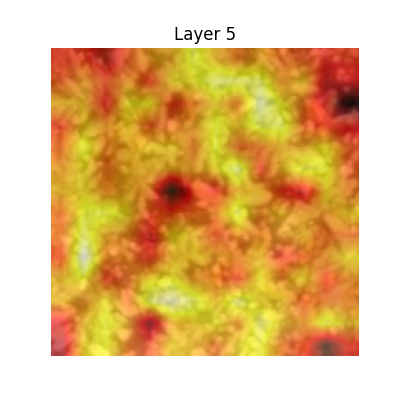}
        \caption{}
        \label{fig:subfig6}
    \end{subfigure}
    \begin{subfigure}[b]{0.24\linewidth}
        \centering
        \includegraphics[width=\linewidth]{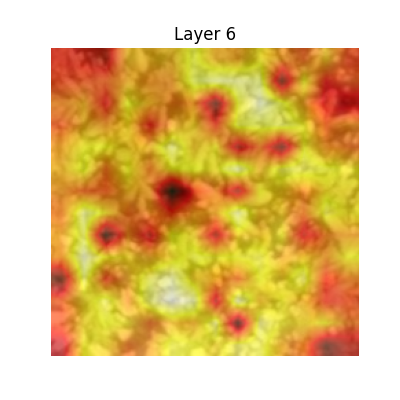}
        \caption{}
        \label{fig:subfig7}
    \end{subfigure}
    \begin{subfigure}[b]{0.24\linewidth}
        \centering
        \includegraphics[width=\linewidth]{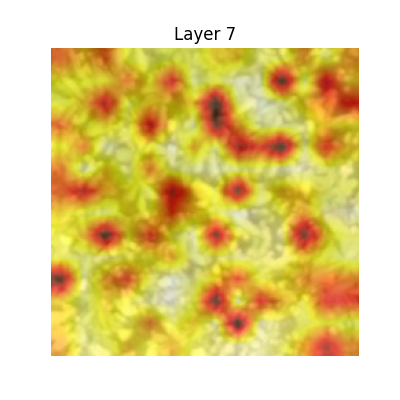}
        \caption{}
        \label{fig:subfig8}
    \end{subfigure}
    \begin{subfigure}[b]{0.24\linewidth}
        \centering
        \includegraphics[width=\linewidth]{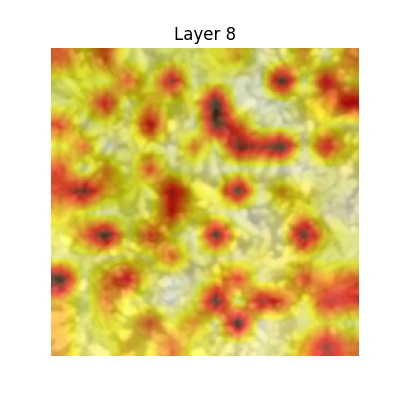}
        \caption{}
        \label{fig:subfig9}
    \end{subfigure}

    \vspace{1em}
    \begin{subfigure}[b]{0.24\linewidth}
        \centering
        \includegraphics[width=\linewidth]{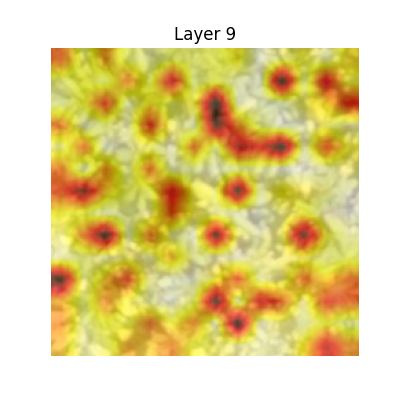}
        \caption{}
        \label{fig:subfig10}
    \end{subfigure}
    \begin{subfigure}[b]{0.24\linewidth}
        \centering
        \includegraphics[width=\linewidth]{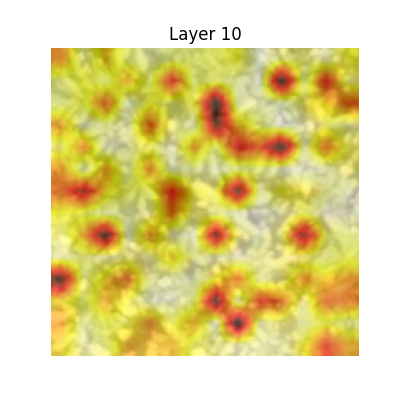}
        \caption{}
        \label{fig:subfig11}
    \end{subfigure}
    \begin{subfigure}[b]{0.24\linewidth}
        \centering
        \includegraphics[width=\linewidth]{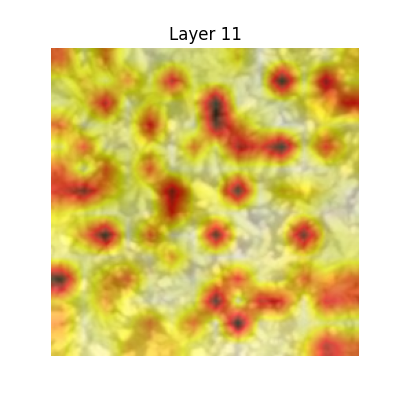}
        \caption{}
        \label{fig:subfig12}
    \end{subfigure}
    \begin{subfigure}[b]{0.24\linewidth}
        \centering
        \includegraphics[width=\linewidth]{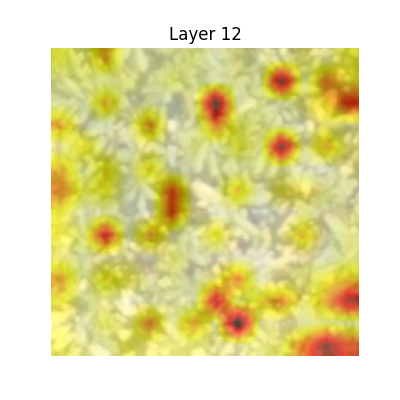}
        \caption{}
        \label{fig:subfig13}
    \end{subfigure}
    
    \caption{A Sample Image (Stressed) and Corresponding Attention Maps from 12 Encoder Blocks.}
    \label{fig:main}
\end{figure}

The stressed image along with the corresponding attention maps from the 12 encoder blocks of the Vision Transformer (ViT) is shown in Fig \ref{fig:main}. Several key observations can be made from these attention maps, including spatial relevance, hierarchical processing, interpreting model decisions, visualization of learned features, and using them as a diagnostic tool for model improvement. Each attention map is resized and overlaid on the original image, with colors indicating the attention intensity:
\begin{itemize}
    \item Red/Hot Colors: Indicate areas of high attention.
    \item Yellow/Warmer Colors: Show areas of moderate  attention.
    \item Dark/Cold Colors: Represent areas of low attention.
\end{itemize}    

\textbf{Spatial Relevance}: The spatial relevance can give insights into which parts of the image the model finds important for differentiating between classes. In Layer 1, the model's attention is broadly distributed with some central intensity, indicating that the model initially captures coarse, global structures of the image. Moving to Layers 2 ,3 and 4, attention becomes increasingly localized, suggesting the model is beginning to identify distinct regions and features relevant to the classification task. Between Layers 5 and 6, the attention patterns grow sharper and more discriminative. These layers appear to focus on intermediate-level features—regions that are potentially indicative of stress patterns but not yet fine-grained. Layers 7 through 11 show a high concentration of attention in specific, small regions with strong contrast, indicating that the model is now attending to fine details, such as localized stress indicators in the vegetation. Notably, Layer 12 diverges from the preceding layers. The attention becomes more diffused and less sharply defined compared to Layers 8–11. This indicates that the final encoder block form a more holistic representation by aggregating information from previous layers,
striking a baalnce of local details with global context for the final classification decision.

\textbf{Hierarchical Processing}: As we move through the layers of the ViT, attention maps can show how the model progressively refines its understanding of the image. In the lower layers, attention is distributed across large regions, capturing global context. As we ascend through the layers, the attention narrows down to more specific features, highlighting finer details and important objects within the image. This hierarchical processing is crucial for the model to effectively balance global and local information.

\textbf{Interpreting Model Decisions}:
By visualizing attention maps, we can interpret why the model made certain predictions. For instance, if the attention maps highlight specific objects or patterns in an image, it suggests that those elements influenced the classification decision. This interpretability can help validate the model’s decisions and identify potential biases or weaknesses.

\textbf{Visualization of Learned Features}:
The attention maps provide a visual representation of the features learned by the ViT. Unlike abstract feature vectors, these maps directly relate model activations to spatial locations in the input image. This visualization helps in understanding how the model processes visual information and forms its internal representations.

\textbf{Diagnostic Tool for Model Improvement}:
Analyzing attention maps can serve as a diagnostic tool for improving model performance. By examining where the model attends and comparing it with ground truth or human perception, we can identify areas where the model might be lacking or where it might over-emphasize certain features. This feedback loop can guide model refinement and training strategies.

In summary, attention maps in image classification tasks with Vision Transformers provide a transparent view into the inner workings of the model, highlighting which parts of the input image contribute most to its decision-making process. 
\subsection{Performance of ViT+SVM}
We used a pre-trained Vision Transformer (ViT) model to extract features from both the training and testing datasets. These extracted features were then utilized by an SVM to identify stress, aiming to distinguish between stressed and healthy images. The implementation was done using the \textit{PyTorch} framework, leveraging both its core library and the \textit{torchvision }module. Key libraries such as \textit{torch}, \textit{torchvision}, \textit{csv}, \textit{pandas}, and \textit{scikit-learn} were imported to facilitate feature extraction, data transformations, file handling, and classification tasks.

We employed the Vision Transformer (\textit{ViT}) models for feature extraction and subsequent evaluation through a Support Vector Machine (SVM) classifier. Primarily, we utilized the \textit{ViT-B/16} architecture, sourced from \textit{torchvision.models}, and loaded pre-trained weights using \textit{torch.load} from a specified path. The model was initialized and set to evaluation mode to facilitate inference-based feature extraction. To standardize the inputs, all images were resized to 224$\times$224 pixels and transformed into tensors. Data loaders were constructed for both the training and testing datasets, using a batch size of 32.
Feature extraction was encapsulated in a dedicated function invoked separately for the training and testing data loaders. The resulting features, along with their corresponding labels, were saved to CSV files for downstream processing. This process was executed within a no-gradient context \textit{torch.no\_grad()} to optimize computational efficiency. The hyperparameters and implementation details used for ViT-based feature extraction and classification are summarized in Table~\ref{table:parameters_vit_svm}.

\begin{table}[h!]
\centering
\begin{tabular}{|>{\raggedright}p{4cm}|>{\raggedright}p{4cm}|}
\hline
\multicolumn{2}{|c|}{\textbf{Parameters}} \\
\hline
\textbf{ViT Parameters} & \textbf{SVM Parameters} \\
\hline
Image size for resizing: 224x224 & Learning rate: 0.001 \\
\hline
Batch size for data loaders: 32 & Kernel: radial basis function \\
\hline
 & All other parameters are default of scikit-learn \\
\hline
\end{tabular}
\caption{Parameters in ViT+SVM Framework}
\label{table:parameters_vit_svm}
\end{table}

To analyze the influence of model capacity, we extracted features under two experimental scenarios. In the first, we used the \textit{ViT-B/16} model (final weights from scenario 8). In the second scenario, we utilized \textit{ViT-L/16}, taking into account the weights in scenario 4. In both settings, the extracted features were fed into a Support Vector Machine (SVM) for classification.
Performance evaluation was conducted on specified test set as well as using 5-fold cross-validation to generate the Receiver Operating Characteristic (ROC) curves, compute the mean Area Under the Curve (AUC), and estimate the mean classification accuracy.

\begin{table}[hbt!]
\centering
\caption{Confusion matrix components and test accuracy across for ViT+SVM.}
\label{tab:Confusion_Matrix_vit_svm}
\begin{tabular}{lccccc}
\toprule
Scenario & TP & TN & FP & FN & \shortstack{Test \\ Accuracy} \\
\midrule
ViT-B/16 +SVM & 650 & 353 & 48 & 84 & 0.8837 \\
ViT-L/16 +SVM & 640 & 364 & 37  & 94 & 0.8845 \\
\bottomrule
\end{tabular}
\end{table}

\begin{table}[h!]
\centering
\caption{Mean Accuracy and AUC for k-fold cross validation}\label{tab1:overall}
\begin{tabular}{@{}lccc@{}}
\toprule
Approaches & \multicolumn{1}{c}{Accuracy for} & \multicolumn{1}{c}{Mean Accuracy} & \multicolumn{1}{c}{Mean AUC} \\
 & \multicolumn{1}{c}{Specified Test Set} & \multicolumn{1}{c}{with 5-Fold} & \multicolumn{1}{c}{with 5-Fold} \\
\midrule
ViT-B/16 +SVM  & 0.8837 & 0.9435 & 0.98 \\
ViT-L/16 +SVM  & 0.8845 & 0.9208 & 0.96 \\
\bottomrule
\label{table:k_fold}
\end{tabular}
\end{table}

\begin{figure}[hbt!]
    \centering
    \label{fig:roc_curves}
    \begin{subfigure}{0.45\textwidth}
    \centering
        \includegraphics[width=0.9\textwidth]{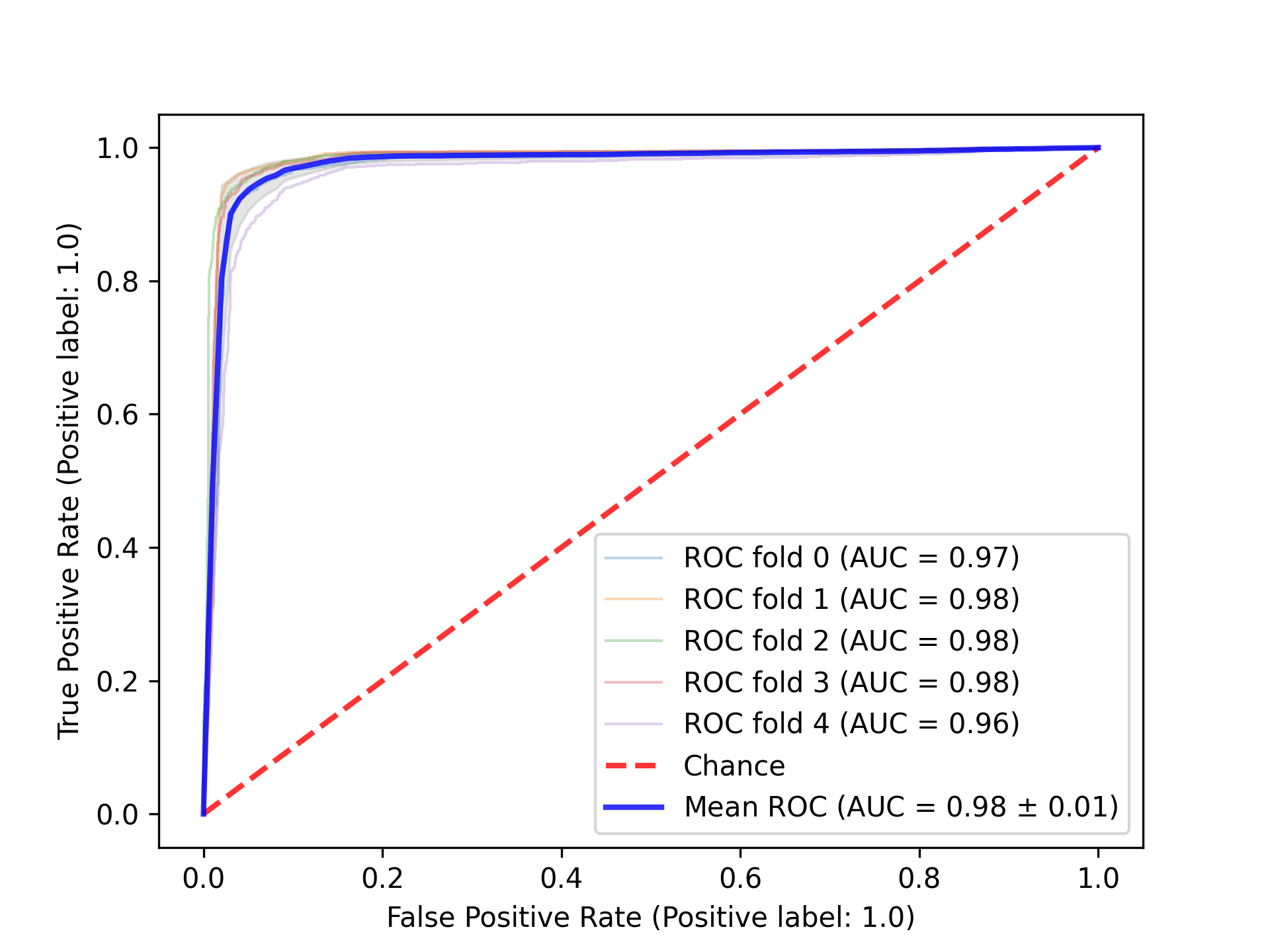}
        \caption{ViT-B/16 +SVM}
        \label{fig:b16}
    \end{subfigure}
    \begin{subfigure}{0.45\textwidth}
    \centering
        \includegraphics[width=0.9\textwidth]{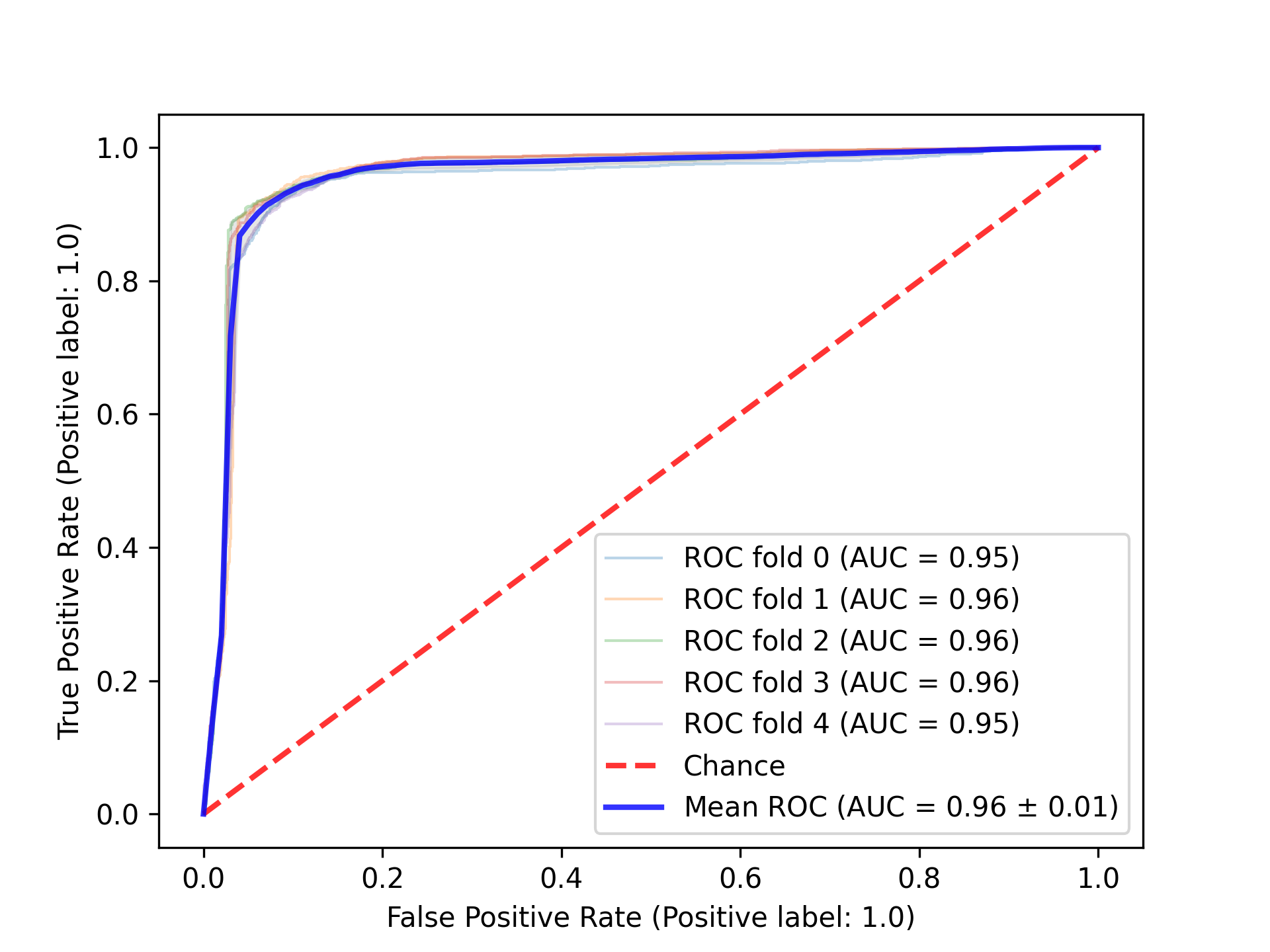}
        \caption{ViT-L/16 +SVM}
        \label{fig:l16}
    \end{subfigure}
    \caption{ROC curves depicting the model's performance}
\end{figure}

The confusion matrix components, as shown in Table \ref{tab:Confusion_Matrix_vit_svm} , show that ViT-B/16 produces a lower number of false negatives (FN), a critical metric in drought stress identification tasks, where missing stressed cases can lead to substantial consequences. The overall performance evaluation of the proposed models reveals the superiority of the ViT-B/16 + SVM architecture over its larger counterpart, ViT-L/16 + SVM. Table \ref{table:k_fold}  detailed that ViT-B/16 consistently achieved higher mean accuracy, mean AUC, indicating better generalization capability and robustness. Additionally, as shown in Fig. \ref{fig:b16}, the ROC curve for ViT-B/16 exhibits a higher mean AUC (0.98 ± 0.01) compared to ViT-L/16 (0.96 ± 0.01 in Fig. \ref{fig:l16}), with consistently strong performance across all folds. The tighter clustering of the ROC curves around the upper-left corner for ViT-B/16 also indicates more reliable classification across varying thresholds. These results demonstrate that despite having a smaller architecture, ViT-B/16 offers more accurate and explainable performance, making it a preferable feature extractor for efficient and dependable drought stress detection.

\subsection{Comparison of the Models}

\textbf{Performance on Specified test set}: 

Our two proposed approaches are compared with a previously published model \cite{patra_explainable_2024} on the same dataset. Table \ref{tab:RGB_Models} presents a comparative evaluation of the three models for drought stress detection, reporting precision, recall, F1-score, and overall accuracy on the designated test set. The CNN-based framework demonstrates competitive performance, particularly in terms of stressed plant precision (0.9673). However, it lags behind in F1-score and overall accuracy when compared to the Vision Transformer with Transfer Learning. The ViT+SVM model shows relatively weaker performance, especially in classifying healthy plants, suggesting that the SVM integration may not fully exploit the representational capacity of ViTs for this task. The Vision Transformer with Transfer Learning outperforms both the CNN-based and ViT+SVM models across all metrics, making it the most effective model for distinguishing drought-stressed from healthy plants.

\begin{figure}[htbp]
    \centering
    
    \begin{subfigure}[b]{0.25\linewidth}
        \centering
        \includegraphics[width=\linewidth]{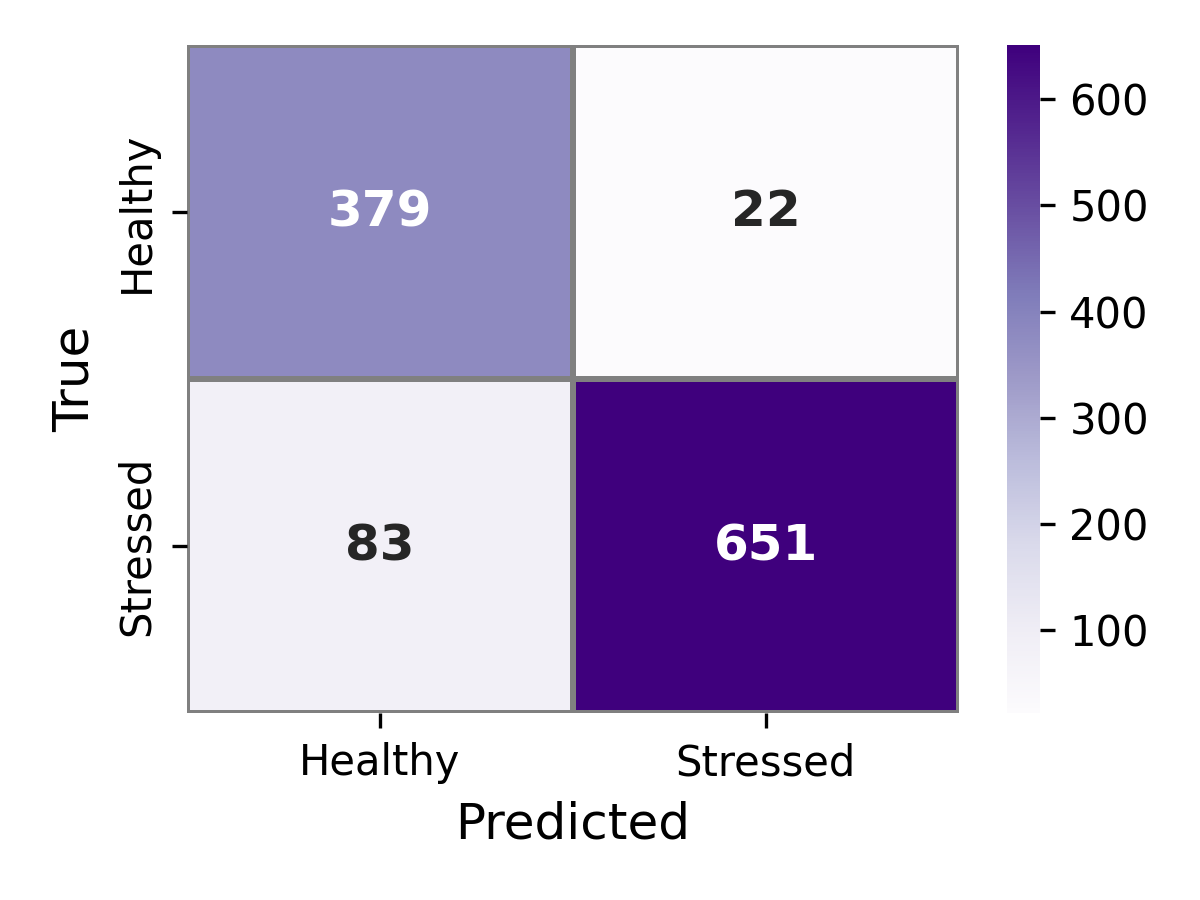}
        \caption{CNN Framework \cite{patra_explainable_2024}}
        \label{fig:conf1}
    \end{subfigure}
    \hfill
    \begin{subfigure}[b]{0.25\linewidth}
        \centering
        \includegraphics[width=\linewidth]{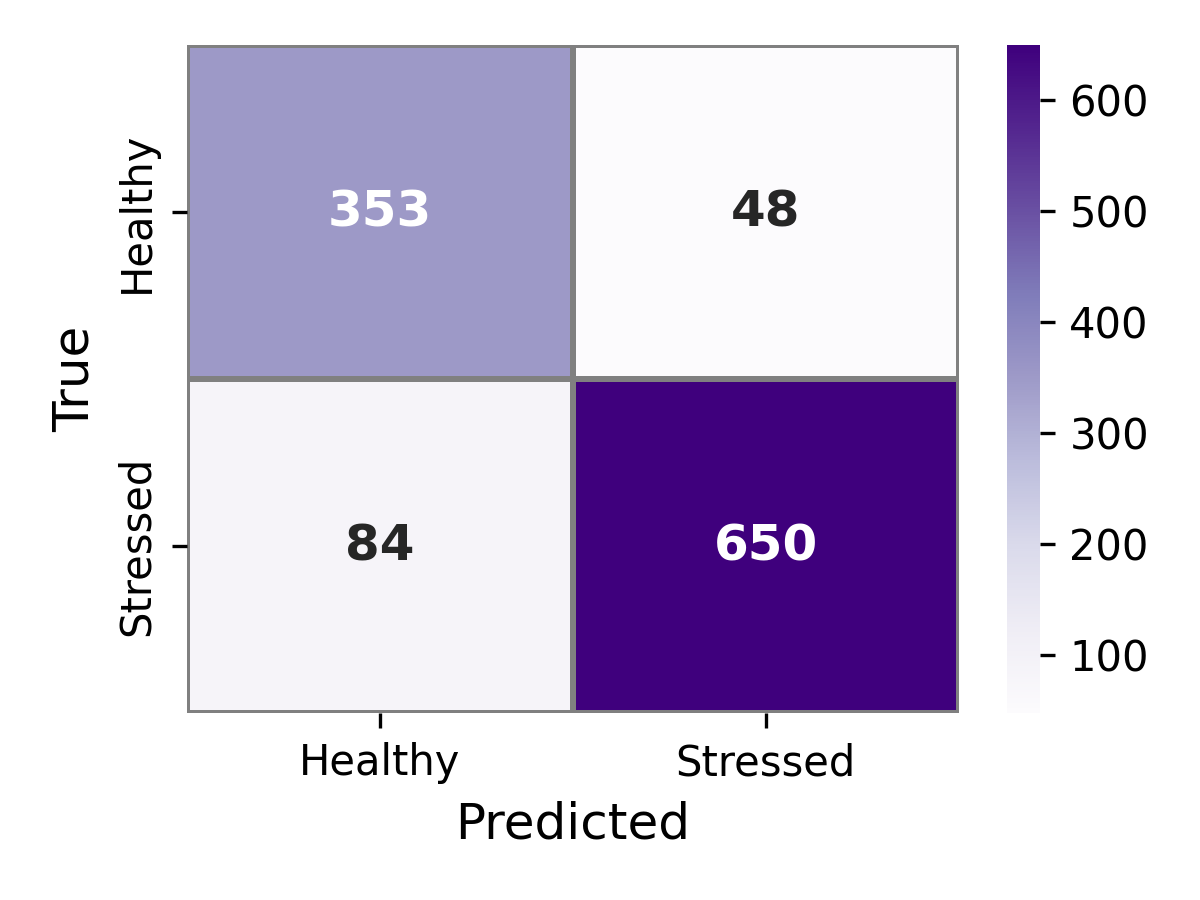}
        \caption{ViT+SVM}
        \label{fig:conf2}
    \end{subfigure}
    \hfill
    \begin{subfigure}[b]{0.25\linewidth}
        \centering
        \includegraphics[width=\linewidth]{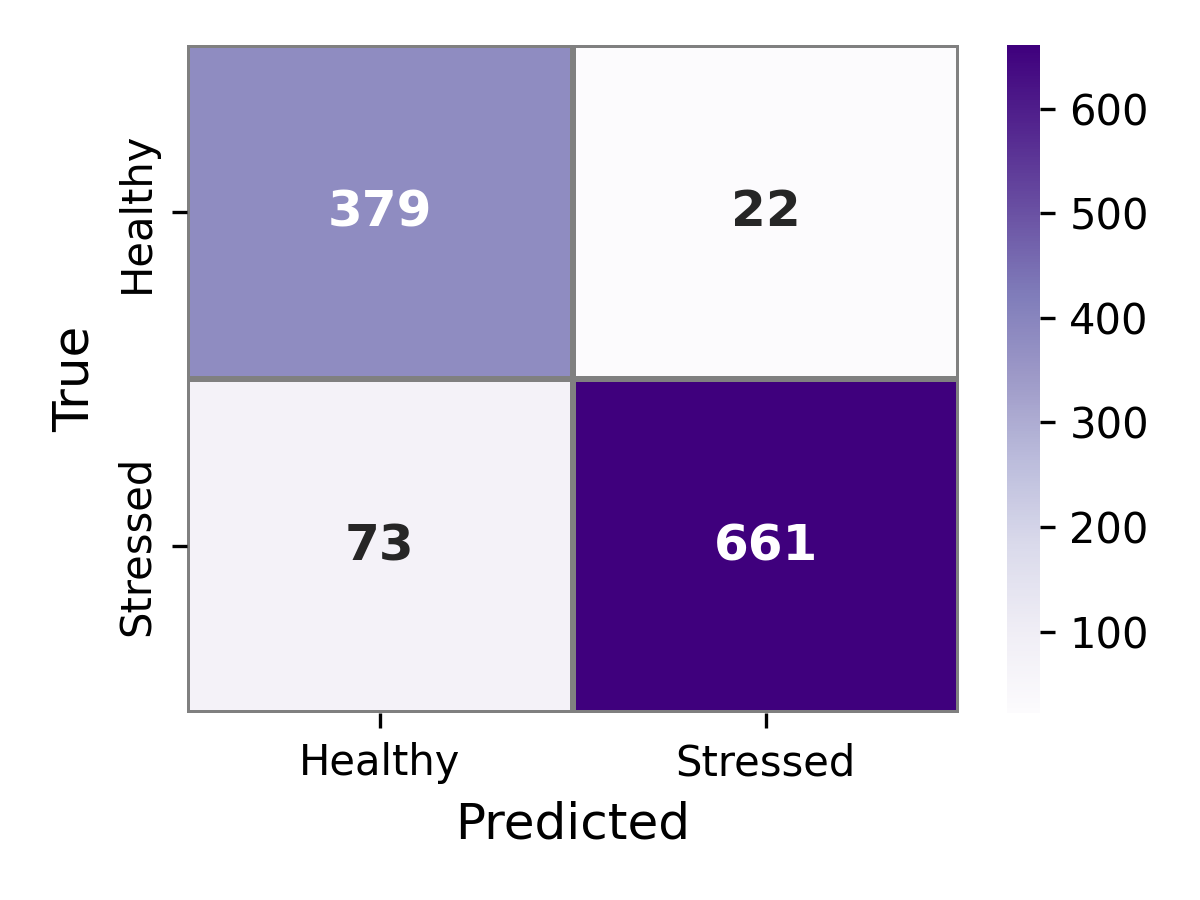}
        \caption{ViT-TL}
        \label{fig:conf3}
    \end{subfigure}
    
    \caption{Confusion matrices comparison for CNN, ViT+SVM, and ViT with Transfer Learning models.}
    \label{fig:con_mat}
\end{figure}

Fig. \ref{fig:con_mat} presents the confusion matrices for three models—CNN Framework, ViT+SVM, and ViT with Transfer Learning—which correspond to the performance metrics previously computed and detailed in Table \ref{tab:RGB_Models} for the specified test set. In drought stress detection, minimizing false negatives (FN) is crucial, as higher FN could lead to significant oversight in applications like drought stress detection, where failing to identify stressed plants might delay critical interventions. Among the three models compared, the Vision Transformer with Transfer Learning (ViT-TL) demonstrates the lowest number of false negatives (73), highlighting its superior sensitivity to actual stress conditions. In contrast, the CNN-based framework and the ViT+SVM model yield higher false negatives—83 and 84 respectively—indicating a comparatively weaker ability to detect all truly stressed plants.
Apparently, false positives (FP) are generally less severe than false negatives, they still represent inefficiencies in resource utilization, such as unnecessary irrigation or pesticide application to healthy plants. Both ViT-TL and the CNN framework exhibit equally low false positives (22), whereas the ViT+SVM model produces a significantly higher FP count of 48, suggesting an over-prediction of stress in healthy plants.
Overall, ViT-TL achieves a well-balanced performance with the lowest FN and equally low FP, indicating its robustness in both detecting stressed plants accurately and avoiding unwarranted false alarms.

\begin{table}[hbt!]
\centering
\caption{Performance comparison of the models for the specified test set.}
\label{tab:RGB_Models}
\begin{tabular}{lccccccc}
\toprule
\multirow{2}{*}{Model} & \multicolumn{3}{c}{Stressed} & \multicolumn{3}{c}{Healthy} & \multirow{2}{*}{Accuracy} \\
\cmidrule(lr){2-4} \cmidrule(lr){5-7}
 & Precision & Recall & F1-score & Precision & Recall & F1-score & \\
\midrule
CNN Based Framework \cite{patra_explainable_2024} & 0.9673 & 0.8869 & 0.9252 & 0.8203 & 0.9451 & 0.8788 & 0.9075 \\ 
ViT+SVM & 0.9312 & 0.8856 & 0.9078 & 0.8078 & 0.8803 & 0.8421 & 0.8837 \\
ViT-TL & \textbf{0.9678} & \textbf{0.9005} & \textbf{0.9328} & \textbf{0.8385} & \textbf{0.9451} & \textbf{0.8883} & \textbf{0.9162} \\
\bottomrule
\end{tabular}
\end{table}

\textbf{K-fold cross validation:}
Table~\ref{tab:ViT_Models} presents a comparative performance analysis between two model configurations: the Vision Transformer (ViT) with transfer learning and ViT-B/16 combined with a Support Vector Machine (SVM) classifier. In the first approach, the ViT model is fine-tuned end-to-end on the target dataset, whereas in the second, ViT is used as a feature extractor, and an SVM classifier is trained on the extracted features. The evaluation was performed using k-fold cross-validation, and the results are reported in terms of the mean F1-scores for both the Healthy and Stressed classes, along with the overall mean accuracy. The ViT with transfer learning demonstrated superior performance with F1-scores of 0.97 and 0.98 for the Healthy and Stressed classes, respectively, and a mean accuracy of 97.43\%. In comparison, the ViT-B/16 + SVM configuration achieved F1-scores of 0.93 (Healthy) and 0.95 (Stressed), and a mean accuracy of 94.35\%. These results highlight the effectiveness of end-to-end fine-tuning, which enables the model to learn task-specific representations more effectively than the frozen feature extractor approach.

\begin{table}[hbt!]
\centering
\caption{Performance comparison between ViT with Transfer Learning and ViT+SVM with Optimal Weights for k-fold cross validation.}
\label{tab:ViT_Models}
\begin{tabular}{lcccc}
\toprule
\multirow{2}{*}{Model} & \multicolumn{2}{c}{Mean F1-score} & \multirow{2}{*}{Mean Accuracy} \\
\cmidrule(lr){2-3}
 & Healthy & Stressed & \\
\midrule
ViT with Transfer Learning & 0.97 & 0.98 & 0.9743 \\ 
ViT-B/16  + SVM & 0.93 & 0.95 & 0.9435 \\
\bottomrule
\end{tabular}
\end{table}

\section{Conclusion}
Drought stress represents a severe threat to crop yield and quality, disrupting normal plant growth and survival rates. Detecting early signs of drought stress is crucial for effective crop management and intervention. Traditional methods, primarily reliant on Convolutional Neural Networks (CNNs), have made significant strides in capturing spatial hierarchies in image data. However, Vision Transformers (ViTs) offer a compelling alternative by leveraging self-attention mechanisms to capture long-range dependencies and complex spatial relationships, thus enhancing the detection of subtle drought stress indicators.

Our study successfully addressed the challenge of stress identification using smaller datasets by harnessing the feature extraction capabilities of Vision Transformers. Unlike conventional CNN architectures, Vision Transformers utilize self-attention mechanisms to effectively capture relationships across different parts of the image, enabling them to model long-range dependencies, which is particularly advantageous for complex datasets where traditional CNNs may struggle to capture global context. Moreover, Vision Transformers can handle images of varying resolutions without necessitating architectural modifications, enhancing their flexibility in handling diverse datasets. Our framework also emphasized explainability by generating attention maps, which provide insights into the model’s focus areas within the images, thereby offering transparency in its decision-making process.

The comparative analysis of a CNN-based framework, ViT-TL, and ViT+SVM showed that Vision Transformers (ViT) with transfer learning significantly improve classification performance. Its performance is consistent across various data splits, thus offering a reliable and accurate solution for drought stress identification, with attention maps providing valuable insights into the model's decision-making process. These findings highlight the potential of advanced deep learning techniques to enhance agricultural practices and decision-making, paving the way for more effective crop management strategies.

\section*{Competing Interests}
The authors have no relevant financial or non-financial interests to disclose.

\section*{Author Contributions}
Aswini Kumar Patra contributed to the study conception and design. He was also responsible for material preparation, data collection, and analysis. The first draft of the manuscript was written by Aswini Kumar Patra, and all authors reviewed and provided feedback on previous versions. Ankit Varshney contributed to the conceptualization of the ViT and SVM combination. Lingaraj Sahoo supervised the research and edited the manuscript. All authors read and approved the final version of the manuscript.

\section*{Data Availability}
The data used in this study are derived from a publicly available dataset, which can be accessed at \url{https://www.webpages.uidaho.edu/vakanski/Multispectral\_Images\_Dataset.html}. Additionally, the generated data supporting the findings of this study are available from the corresponding author upon reasonable request.

\bibliographystyle{unsrt}
\bibliography{sn-bibliography1}

\end{document}